\documentclass{article}

    \PassOptionsToPackage{numbers, compress}{natbib}

\usepackage[preprint]{neurips_2024}

\usepackage[utf8]{inputenc} 
\usepackage[T1]{fontenc}    
\usepackage{hyperref}       
\usepackage{url}            
\usepackage{booktabs}       
\usepackage{amsfonts}       
\usepackage{nicefrac}       
\usepackage{microtype}      
\usepackage{xcolor}         

\usepackage{graphicx}
\usepackage{amsmath}
\usepackage{amssymb}
\usepackage{overpic}
\usepackage{tabularx}
\usepackage{multirow}
\usepackage{tabularray}
\usepackage{orcidlink}
\usepackage{subcaption}
\usepackage{wrapfig}

\newcommand{\R}{\mathbb{R}}
\let \bs=\mathbf
\let \set=\mathcal

\def \data {\textup{data}}

\def \path {\mathit{path}}

\def \topo {\textup{topo}}
\def \regu {\textup{regu}}

\def \deform {\textup{def}}

\def \fit {\textup{fit}}
\def \arap {\textup{arap}}
\def \arap {\textup{acap}}

\def \sreg {\textup{sr}}
\def \soff {\textup{so}}
\def \creg {\textup{cr}}
\def \sEDR {\textup{se}}

\def \arap {\textup{arap}}

\def \exp {\textup{exp}}

\let \set = \mathcal
\let \bs = \boldsymbol

\title{4DRecons: 4D Neural Implicit Deformable Objects Reconstruction from a single RGB-D Camera with Geometrical and Topological Regularizations}

\author{%
  Xiaoyan Cong\thanks{Work done while an intern at The University of Texas at Austin.} \\
  Zhejiang University\\
  \And
  Haitao Yang \\
  The University of Texas at Austin \\
  \And
  Liyan Chen \\
  The University of Texas at Austin \\
  \And
  Kaifeng Zhang \\
   University of Illinois Urbana-Champaign \\
  \And
  Li Yi \\
  Tsinghua University \\
  \And
  Chandrajit Bajaj \\
  The University of Texas at Austin  \\
  \And
  Qixing Huang\thanks{Corresponding author.} \\
  The University of Texas at Austin  \\
}

\begin{document}

\maketitle


\begin{abstract}
This paper presents a novel approach 4DRecons that takes a single camera RGB-D sequence of a dynamic subject as input and outputs a complete textured deforming 3D model over time. 4DRecons encodes the output as a 4D neural implicit surface and presents an optimization procedure that combines a data term and two regularization terms. The data term fits the 4D implicit surface to the input partial observations. We address fundamental challenges in fitting a complete implicit surface to partial observations. The first regularization term enforces that the deformation among adjacent frames is as rigid as possible (ARAP). To this end, we introduce a novel approach to compute correspondences between adjacent textured implicit surfaces, which are used to define the ARAP regularization term. The second regularization term enforces that the topology of the underlying object remains fixed over time. This regularization is critical for avoiding self-intersections that are typical in implicit-based reconstructions. We have evaluated the performance of 4DRecons on a variety of datasets. Experimental results show that 4DRecons can handle large deformations and complex inter-part interactions and outperform state-of-the-art approaches considerably. 
\end{abstract}

\section{Introduction}

We are interested in reconstructing a deforming object from a single RGB-D sensor. This problem has been extensively studied in the literature. Early approaches~\cite{DBLP:conf/sgp/WandJHBGS07,DBLP:journals/tog/WandAOBBJGSS09,DBLP:conf/cvpr/NewcombeFS15} are based on aligning and merging frames. However, errors can accumulate and it is difficult to address penetrations and self-intersections. Recent approaches~\cite{bozic2020deepdeform,Cai2022NDR,lin2022occlusionfusion} leverage deep neural networks to perform non-rigid registration and fusion. Still, inter-penetrations and self-intersections remain a glaring issue. 

In this paper, we propose to formulate 4D dynamic reconstruction as learning a 4D implicit field problem (iso-value of surface and colors) from partial RGB-D scans. This is motivated by the success of implicit neural representations in representing and encoding static~\cite{Park_2019_CVPR,https://doi.org/10.48550/arxiv.2210.00379} and dynamic~\cite{DBLP:conf/cvpr/PumarolaCPM21,DBLP:conf/iccv/ParkSBBGSM21,DBLP:conf/iccv/TretschkTGZLT21} objects and scenes. Our approach 4DRecons combines a data term and two regularization terms. The data term fits the implicit field to the input scans. We introduce an approach to define the data term so that the data term is well defined even when the observation is partial. The first regularization term regularizes the deformation between adjacent frames and the smoothness of the deformations among triplets of frames. This is achieved by a novel approach that computes correspondences between implicit fields, using which we define the regularization terms. 

The second regularization term, which is a key contribution of this paper, enforces that the topology of the reconstruction remains fixed over time. This constraint, combined with the implicit field representation, nicely addresses the open problem of obtaining self-intersection-free reconstructions under the explicit representation, e.g., deforming the SMPL model. Our approach is based on recent advances in the optimization of geometry with given topological constraints~\cite{DBLP:journals/cgf/GabrielssonGSG20,DBLP:journals/cgf/PoulenardSO18,DBLP:conf/cvpr/MezghanniBLO21} and is easy to optimize.

\begin{figure*}[t]
\centering
\includegraphics[width=0.9\linewidth]{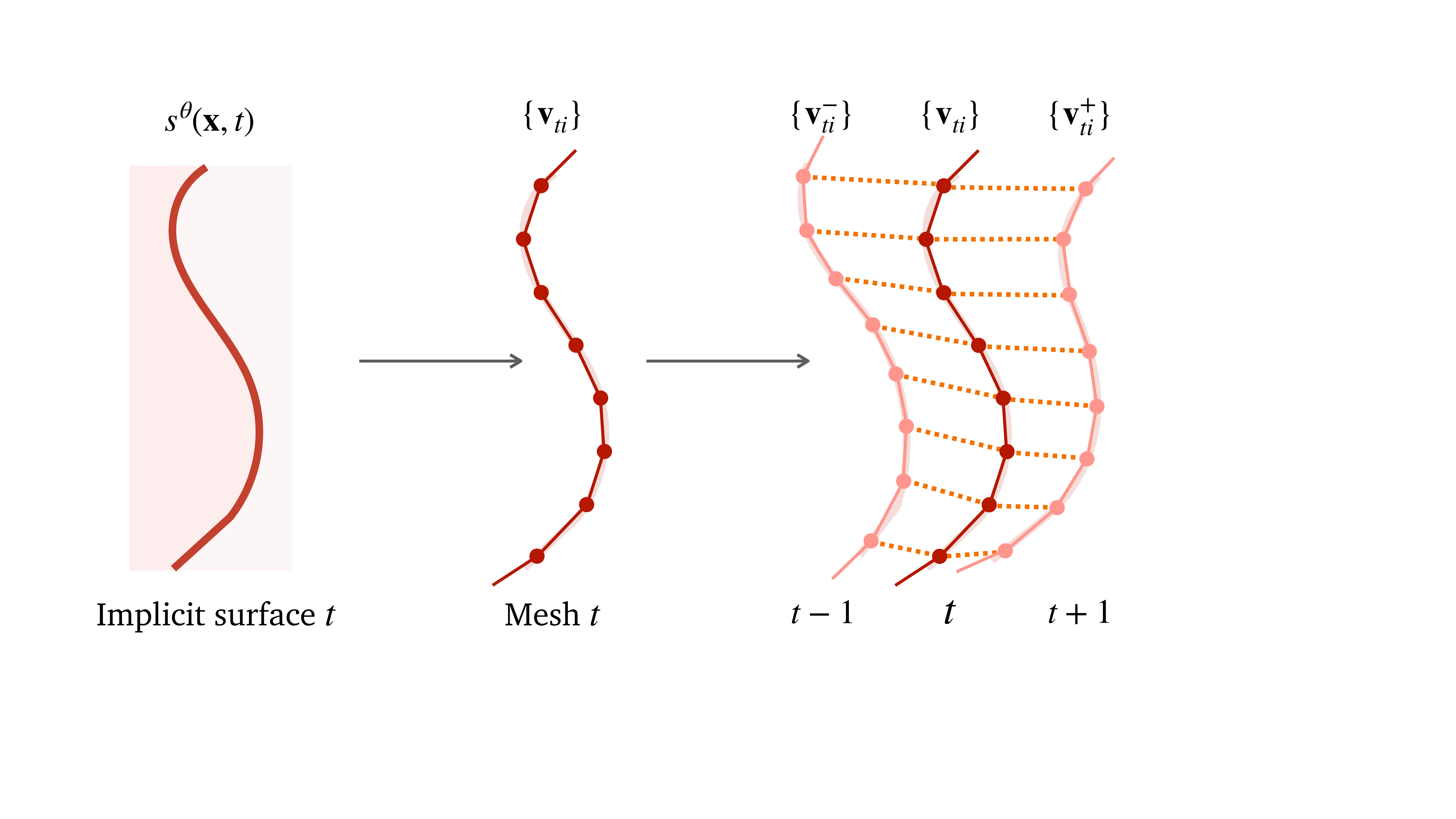}
\caption{Pipeline overview of 4DRecons, which performs a four-stage optimization procedure. The first stage initializes the geometry field by fitting the input data. The second stage enforces deformation and topology regularizations to improve the geometry field. The third stage initializes the color field while fixing the geometry field. The last stage jointly refines the geometry field and the color field. }
\label{Figure:Overview}
\vspace{-0.15in}
\end{figure*}


We have evaluated 4DRecons on a variety of datasets. The experimental results show that 4DRecons can handle large deformations and complex inter-part interactions. Quantitatively, 4DRecons considerably outperforms state-of-the-art dynamic reconstructions. 
\section{Related Works}


\noindent\textbf{Dynamic geometry reconstruction.}
4DRecons falls into the category of animation reconstruction~\cite{DBLP:conf/sgp/WandJHBGS07,DBLP:journals/tog/WandAOBBJGSS09}, which has been studied in graphics and 3D vision for more than two decades. Animation reconstruction aims to recover the complete 3D model and the underlying deformations from an RGB-D scan sequence, where each frame captures a deformation object from one view. This problem is generalized from the rigid object reconstruction problem from depth scans, which has a long literature. A fundamental challenge of animation reconstruction is to recover inter-frame correspondences for aggregating information from multiple frames into a complete model. Many animation reconstruction approaches require a template model, but they do not apply in our setting, as we do not assume that we have a template model as an input. 

Template-free approaches fall into explicit methods and implicit methods. Early works are explicit methods (e.g., DynamicFusion~\cite{DBLP:conf/cvpr/NewcombeFS15}) that progressively align the next frame to the current reconstruction that aggregates all existing frames. Deformations are modeled using deformation graphs~\cite{DBLP:journals/tog/SumnerSP07,DBLP:journals/cgf/HuangAWG08,DBLP:journals/cgf/LiSP08,DBLP:journals/tog/LiAGP09} or volumetric deformations~\cite{DBLP:conf/cvpr/NewcombeFS15}. One limitation of explicit methods is that it is very difficult to handle self-collision. This issue is addressed by implicit methods, which seek to reconstruct time-varying implicit surfaces. In the literature, people have studied different ways to regularize implicit surfaces. Sharf et al.~\cite{DBLP:journals/tog/SharfALGSAC08} and \cite{DBLP:conf/iccv/TretschkTGZLT21} studied how to enforce the imcompressivity of a deforming object in the reconstruction procedure. KillingFusion~\cite{DBLP:conf/cvpr/SlavchevaBCI17} studied how to model the local rigidity constraint by borrowing ideas from Killing vector fields. 

4DRecons falls into the category of implicit methods and presents two fundamental contributions. First, deformation is modeled on the underlying surface, in contrast to the volumetric field employed in KillingFusion. This approach places less constraint on the underlying volumetric field and is more flexible. Second, we enforce topological consistency across the input frames, avoiding merging contacting surfaces under the implicit representation.

\noindent\textbf{Neural implicits from point clouds.}
DeepSDF~\cite{Park_2019_CVPR} is pioneering the research area of reconstructing an implicit surface from point-cloud data. The key idea is to generate samples inside and outside the surface, which are used to regress a volumetric neural implicit field. Several approaches improved DeepSDF performance by using better data losses, e.g., SAL~\cite{DBLP:conf/cvpr/AtzmonL20} and SALD~\cite{DBLP:conf/iclr/AtzmonL21}. However, these formulations cannot handle partial observations because the samples generated near the boundary area are not well-defined. 4DRecons uses a different approach which carefully places samples close to the observed surface area and constrains the signed distance function as a loss term on its gradient field. 
In particular, our sampling strategy, which is based on an analysis of the confidence of the sample, is critical to ensure a high-quality implicit surface. 

Another way to address partial observations is to reconstruct an unsigned distance function (UDF) from the observed points~\cite{DBLP:conf/nips/ChibaneMP20,DBLP:conf/nips/ZhouMLFH22,Liu_2023_CVPR,Long_2023_CVPR}. However, a fundamental challenge of UDF is to extract the underlying surface. Moreover, it is applied mainly to objects with boundaries, and the observation in each frame is complete~\cite{Long_2023_CVPR}. The reason is that under UDF it is very difficult to aggregate partial observations at different frames to form a complete surface. 

\noindent\textbf{Deformation modeling.}
Embedded deformations~\cite{10.1145/1276377.1276478} are widely used for dynamic reconstruction. The technical challenge in our setting is that embedded deformations require an explicit geometric representation, which is not available in our setting. 4DRecons innovates in computing dense correspondences between adjacent implicit surfaces. Unlike GenCorres~\cite{DBLP:journals/corr/abs-2304-10523} where correspondences are completely driven by geometry, 4DRecons computes correspondences by matching geometry and color. Using these correspondences, we introduce regularization terms that penalize deformations and enforce color consistency. 

Many dynamic reconstruction approaches penalize the deformation between a template model and the input scans. However, this approach requires either a template model~\cite{DBLP:conf/iccv/NoguchiSLH21,DBLP:conf/cvpr/PumarolaCPM21,DBLP:journals/tog/LiuHRSGT21}, which is not always available, or treats the first frame as a template model~\cite{DBLP:conf/iccv/ParkSBBGSM21}, which requires the first frame to be complete. Moreover, these approaches cannot handle large deformations, which are difficult to model. In contrast, 4DRecons minimizes deformations between adjacent frames. The deformations between non-adjacent frames can still be large. 


\noindent\textbf{Topological regularization.}
4DRecons is motivated by recent work on optimizing a 3D shape with prescribed topology~\cite{DBLP:journals/cgf/GabrielssonGSG20,DBLP:journals/cgf/PoulenardSO18} and enforcing that the shape generator outputs connected 3D shapes~\cite{DBLP:conf/cvpr/MezghanniBLO21}. The basic idea is to link the vertices of a 3D shape with topological features on the persistent diagram~\cite{books/daglib/0025666}. This allows us to deform a 3D shape to match topological attributes. 4DRecons enforces that the number of topological features on the persistent diagram remains fixed across all the frames. This approach nicely penalizes inter-penetrations that frequently exhibit in explicit-based and implicit-based dynamic reconstructions. To the best of our knowledge, 4DRecons is the first approach that enforces topological consistency for dynamic reconstruction. 


\begin{figure*}[t]
\centering
\includegraphics[width=0.9\linewidth]{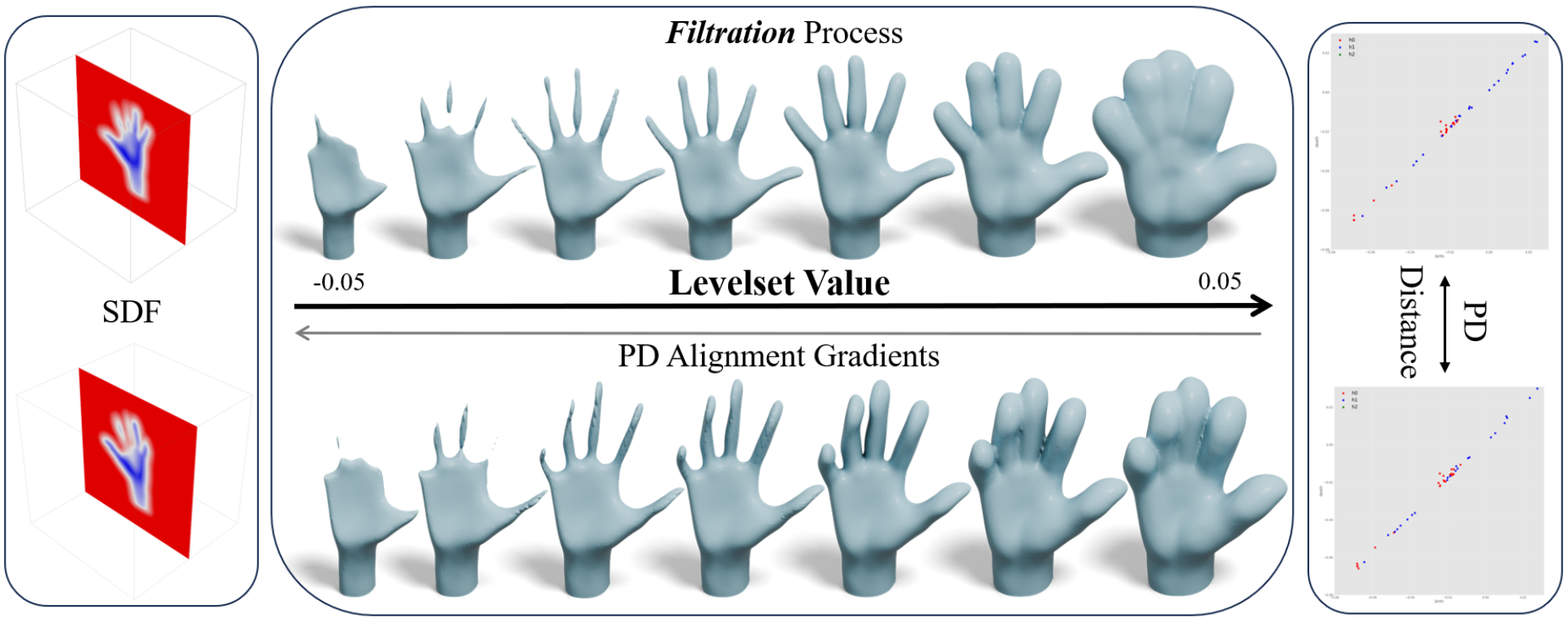}
\caption{The \textit{filtration} processes of two SDF volumes, resulting in two comparable PD plots that we enforce alignment. Our method allows the alignment gradients to back-propagate and directly manipulate the underlining SDF volumes to synchronize their topological signatures.}
\vspace{-0.2in}
\label{Figure:PD_2}
\end{figure*}

\section{Problem Statement and Approach Overview}

We begin with the problem statement in Section~\ref{Subsec:Problem:Statement}. We then present an overview of our approach in Section~\ref{Subsec:Approach:Overview}.

\subsection{Problem Statement}
\label{Subsec:Problem:Statement}

The input consists of $N$ RGB-D scans $\set{P} \mathrel{:=} \{\set{P}_t | \set{P} = (\bs{x}_{ti},\bs{n}_{ti},\bs{c}_{ti}, t),1\leq t \leq N\}$ represented in the sensor's local coordinate system. Here, $\bs{x}_{ti}\in \R^3$ denotes the position of the sample. $\bs{n}_{ti}\in \R^3$ denotes the normal of the sample. $\bs{n}_{ti}$ is estimated using~\cite{10.1145/777792.777840} and is oriented using the camera center. $\bs{c}_{ti}\in \R^3$ denotes the color of the sample. Our goal is to reconstruct a 4D implicit field $\bs{f}^{\theta}:\R^3 \times \R \rightarrow \R\times \R^3$, which takes a position $\bs{x}$ and time $t$ as input and outputs an isovalue $s$ and color $\bs{c}$, that is, $f^{\theta}(\bs{x},t)\rightarrow (s^{\theta}(\bs{x},t),\bs{c}^{\theta}(\bs{x},t))$. Due to the space constraint, we defer the details of network architecture to the supp. material.

\begin{figure}[tb]
  \centering
  \begin{subfigure}{0.43\linewidth}
    \centering
    \includegraphics[height=2.7cm]{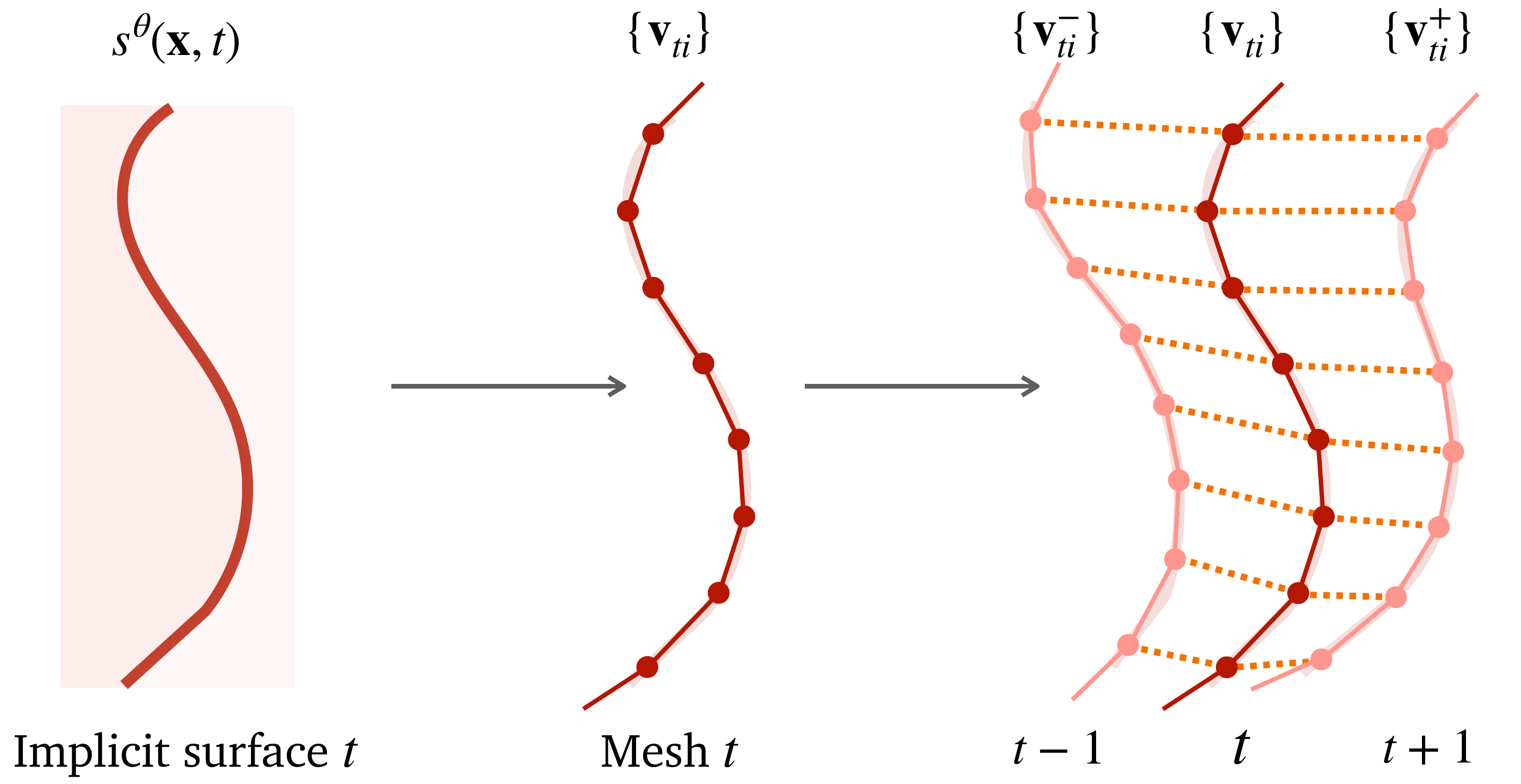}
    \caption{}
    \label{Fig:Implicit:Discretization}
  \end{subfigure}
  \hfill
  \begin{subfigure}{0.56\linewidth}
    \includegraphics[height=2.8cm]{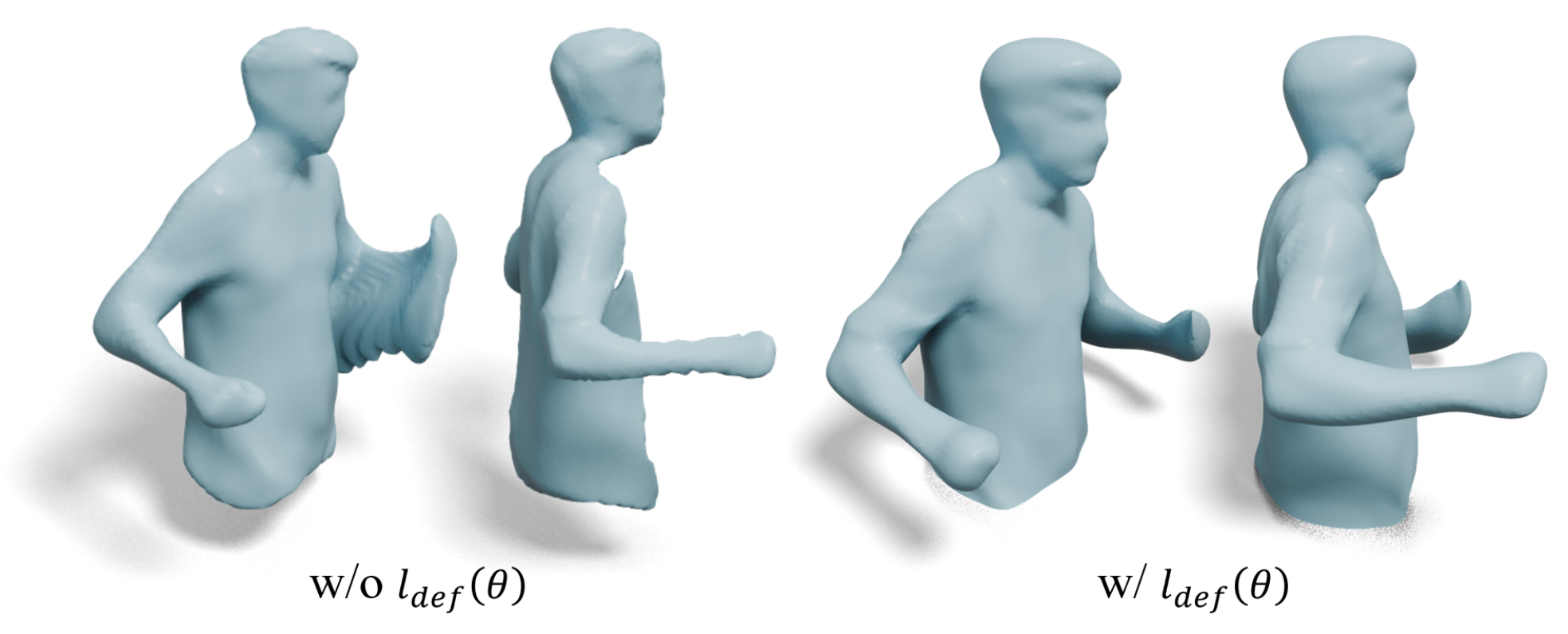}
    \caption{}
    \label{Figure:Ablation:Deformation}
  \end{subfigure}
  \vspace{-0.2in}
  \caption{(a) We first discretize an implicit surface in frame $t$ into a triangular mesh. Then we solve an optimization problem to determine the correspondences of the vertices of this mesh in the next frame $t+1$ and the previous frame $t-1$. These correspondences are then used to define the deformation regularization term. (b) Comparisons of the geometry field reconstruction with (right) and without (left) the deformation regularization term $l_{\deform}(\theta)$ defined in \ref{Subsec:Def:Reg:Term}. }
  \label{fig:short}
\end{figure}

\subsection{Approach Overview}
\label{Subsec:Approach:Overview}

We optimize the network parameters $\theta$ by combining one data loss and two regularization losses:
\begin{equation}
\min\limits_{\theta} l_{\data}(\theta,\{\set{P}\}) + \lambda_{\deform}l_{\deform}(\theta) + \lambda_{\topo}l_{\topo}(\theta)  
\label{Eq:Total:Opt}
\end{equation}
In the following, we highlight the main ideas of our loss terms. Section~\ref{Section:Formulation} explains the technical details. 

\noindent\textbf{Data term.} The data term measures the distance between the implicit field and the input scans. We fit $f^{\theta}$ to samples close to each input scan $\set{P}_t$. We explicitly model scan boundaries to make the SDF constraints well-defined. In addition to data fitting, we also enforce four regularization terms to improve the implicit field in each frame.

\noindent\textbf{Deformation regularization term.} This term regularizes the deformation between two adjacent frames and the smoothness of the deformations between three consecutive frames. Generalizing GenCorres~\cite{DBLP:journals/corr/abs-2304-10523}, we introduce a novel approach to compute the correspondences between textured implicit surfaces. Using these correspondences, we then develop the regularization terms for deformations, deformation smoothness, and color consistency.

\noindent\textbf{Topology regularization term.} The second regularization term enforces that the topology of the implicit reconstruction remains fixed during the optimization procedure. We use the persistent diagram (PD) tool and enforce that the PD of the reconstruction in frame $t$ matches the PD of the reconstruction in frame $t'$, where $(t,t')$ is chosen as a dense subset of frame pairs. 

\section{Approach}
\label{Section:Formulation}

We begin by introducing the data loss in Section~\ref{Subsec:Data:Term}. We then describe the losses in the deformation and topology regularization in Section~\ref{Subsec:Def:Reg:Term} and Section~\ref{Subsec:Top:Reg:Term}, respectively. Finally, we describe the optimization procedure in Section~\ref{Subsec:Optimization}.

\subsection{Data Term}
\label{Subsec:Data:Term}

We define the total data loss as the sum of the loss associated with each scan, which combines a fitting term $l_{\fit}$, and a regularization term $l_{\regu}$.
\begin{equation}
    l_{\data}(\theta,\{\set{P}_t\})  =  \sum\limits_{t=1}^{N}\big(l_{\fit}(\set{P}_t, \bs{f}^{\theta}(\cdot, t)) + l_{\regu}(\set{P}_t, \bs{f}^{\theta}(\cdot, t)))
\label{Eq:Total:Data:Loss}
\end{equation}
In the following, we define each term in detail.

\noindent\textbf{Data fitting term.} We define the data fitting term by generating $m_t$ samples $\Omega_t \mathrel{:=} \{ \bs{p}_{tj} = \big(\bs{x}_{tj},\bs{c}_{tj},\bs{n}_{tj},s_{tj}\big), 1\leq j \leq m_t\}$ from $\set{P}_t$. Here, $\bs{x}_{tj}$ is the sample's position; $\bs{c}_{tj}$ is the sample's color; $\bs{n}_{tj}$ is the sample's normal; $s_{tj}\in R$ is the signed distance value. 

Our sampling strategy is guided by a maximum offset value $d_{\max}$ and an interior confidence value $b_{ti}\in [0,1]$ at $\bs{p}_{ti}$. The interior confidence value $b_{ti}$ is based on projecting the nearest neighbors of $\bs{p}_{ti}$ onto the tangent plane at $\bs{p}_{ti}$, i.e., interior points have neighboring points spread out in all directions. We defer the details to the supp. material as this is not our main contribution. 


We pick a random point with index $i$ and generate a sample as $(\bs{p}_{t}+\psi b_{ti}d_{\max}\bs{n}_{tj}$, $ \bs{c}_{tj}, \psi b_{ti}d_{\max})$ where $\psi\sim \set{N}(0,1)$ is a random variable that follows the normal distribution. This procedure is repeated $M_t$ times. 

Given these samples, we define the fitting term as 
\begin{align}
l_{\fit}(&\set{P}_t, \bs{f}^{\theta}(\cdot, t)) \mathrel{:=}  \int_{\Omega_t} \big((s^{\theta}(\bs{x}_{tj},t)-s_{tj})^2 + \nonumber \\ & \mu_{c}(\bs{c}^{\theta}(\bs{x}_{tj},t)-\bs{c}_{tj})^2 + \mu_{n}(\nabla_{\bs{x}}\bs{s}^{\theta}(\bs{x}_{tj},t)-\bs{n}_{tj})^2\big) d\bs{p}_{tj}
\label{Eq:Fitting:Term}
\end{align}
where $\mu_{c}=0.1, \mu_{n}=0.1$.

\noindent\textbf{Regularization term.}
We define the regularization term $l_{\regu}$ as
\begin{align}
    l_{\regu}(\set{P}_t, \bs{f}^{\theta}(\cdot, t)) = \lambda_{\sreg}\cdot l_{\sreg}(s^{\theta}(\cdot, t)) +\lambda_{\creg}\cdot l_{\creg}(\bs{f}^{\theta}(\cdot, t)) + \lambda_{\soff}\cdot l_{\soff}(s^{\theta}(\cdot, t)) + \lambda_{\sEDR}\cdot l_{\sEDR}(\bs{s}^{\theta}(\cdot, t)) \big)
\label{Eq:DataRegu:Term}
\end{align}
Since the samples are close to the input scan, merely fitting the signed distance, color and normal values of the samples can lead to an erroneous field that is far away from the input scan. To address this issue, we add an Eikonal term $l_{\sreg}(s^{\theta}(\cdot, t))$ to regularize the derivative of the signed distance function~\cite{DBLP:conf/nips/YarivGKL21,DBLP:conf/siggraph/YarivHRVSSBM23,gropp2020implicit}:
\begin{equation}
l_{\sreg}(s^{\theta}(\cdot, t)) \mathrel{:=} \int_{\Omega}\big(\|\nabla s^{\theta}(\bs{x}_{tj},t)\|^2-1\big)^2d\bs{p}_{tj}
\label{Eq:SDF:Regu}    
\end{equation}

Likewise, we define $l_{\creg}(\bs{f}^{\theta}(\cdot, t))$ so that the gradient of the color field is perpendicular to the gradient of the distance field, i.e., color does not change in the normal direction of the underlying surface:
\begin{equation}
l_{\creg}(s^{\theta}(\cdot, t)) \mathrel{:=} \int_{\Omega}\|\nabla s^{\theta}(\bs{x}_{tj},t)^T\nabla \bs{c}^{\theta}(\bs{x}_{tj},t)\|^2d\bs{p}_{tj}
\label{Eq:DCF:Regu}    
\end{equation}

We introduce two additional regularization terms to further improve the quality of the resulting implicit field. Following~\cite{sitzmann2019siren}, we introduce $l_{\soff}(s^{\theta}(\cdot, t))$ to penalize off-surface samples' signed distance value close to $0$:
\begin{equation}
l_{\soff}(s^{\theta}(\cdot, t)) \mathrel{:=} \int_{\Omega_{\textup{off}}}\ \psi(s^{\theta}(\bs{x}_{tj},t)) d\bs{p}_{tj}
\label{Eq:Off:Regu}    
\end{equation}
where $\psi(\bs{x}) = \text{exp}(-\alpha \cdot |x|), \alpha \gg 1$. Furthermore, following~\cite{shue20233d}, we add $l_{\sEDR}(\bs{s}^{\theta}(\cdot, t))$ to enforce the smoothness in the outside-of-shape volume:
\begin{equation}
l_{\sEDR}(\bs{s}^{\theta}(\cdot, t)) \mathrel{:=} \int_{\set{B}}\ (s^{\theta}(\bs{x}_{i},t) - s^{\theta}(\bs{x}_i + \bs{w}_{i},t))^2d\bs{x}_i
\label{Eq:sEDR:Regu}    
\end{equation}
where $\set{B}$ denotes a set of random samples from the outside-of-shape volume; $\bs{w}$ denotes a random vector.


Figure~\ref{Figure:Data:Term} shows that the regularization term $l_{\regu}$ in~\ref{Eq:DataRegu:Term} is important in ensuring the quality of the geometry and color field, particularly near the boundary regions. 
\begin{figure*}[t]
\centering


\begin{minipage}[b]{0.48\textwidth}
\centering
\begin{subfigure}{0.27\textwidth}
\includegraphics[width=\linewidth]{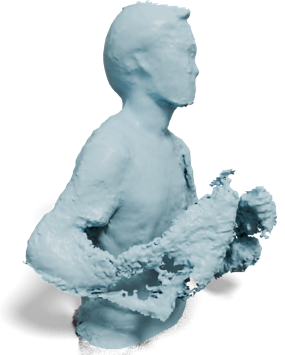}
\end{subfigure}
\hfill
\begin{subfigure}{0.33\textwidth}
\includegraphics[width=\linewidth]{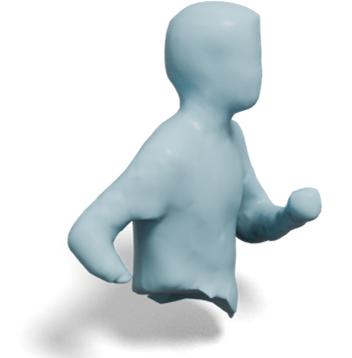}
\end{subfigure}
\hfill
\begin{subfigure}{0.3\textwidth}
\includegraphics[width=\linewidth]{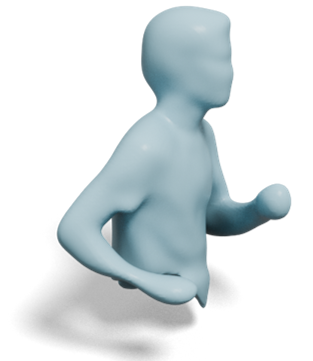}
\end{subfigure}
\par\vspace{0pt} 
\begin{subfigure}{0.31\textwidth}
\includegraphics[width=\linewidth]{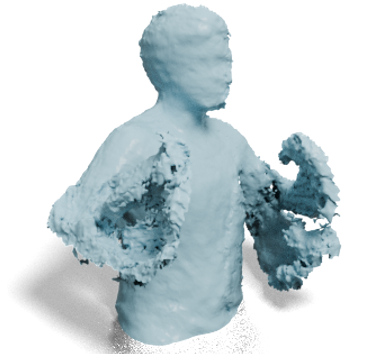}
\caption*{(a)} 
\end{subfigure}
\hfill
\begin{subfigure}{0.3\textwidth}
\includegraphics[width=\linewidth]{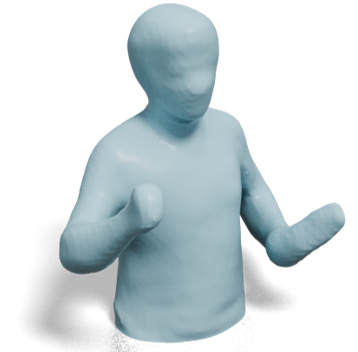}
\caption*{(b)} 
\end{subfigure}
\hfill
\begin{subfigure}{0.29\textwidth}
\includegraphics[width=\linewidth]{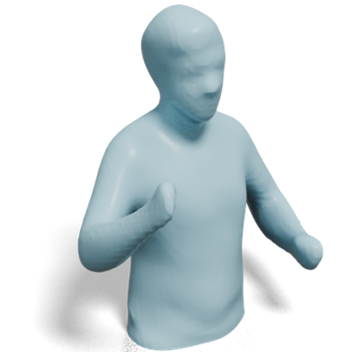}
\caption*{(c)} 
\end{subfigure}
\par\vspace{0pt} 
\end{minipage}
\hfill
\vline\hfill
\begin{minipage}[b]{0.48\textwidth}
\centering

\begin{subfigure}{0.31\textwidth}
\includegraphics[width=\linewidth]{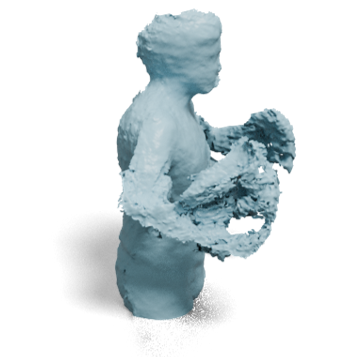}

\end{subfigure}
\hfill
\begin{subfigure}{0.32\textwidth}
\includegraphics[width=\linewidth]{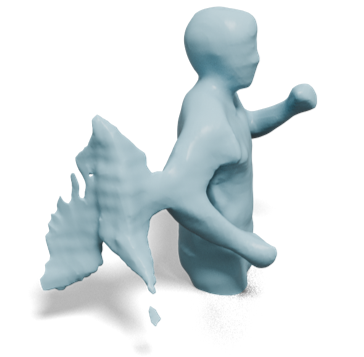}

\end{subfigure}
\hfill
\begin{subfigure}{0.32\textwidth}
\includegraphics[width=\linewidth]{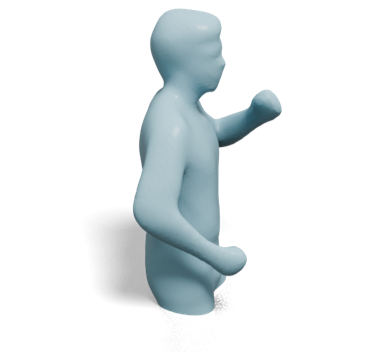}

\end{subfigure}
\par\vspace{0pt} 

\begin{subfigure}{0.32\textwidth}
\includegraphics[width=\linewidth]{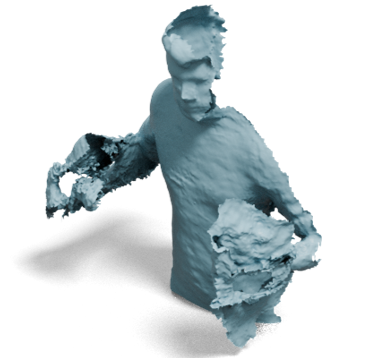}
\caption*{(a)} 
\end{subfigure}
\hfill
\begin{subfigure}{0.3\textwidth}
\includegraphics[width=\linewidth]{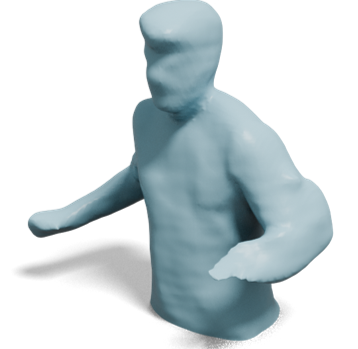}
\caption*{(b)} 
\end{subfigure}
\hfill
\begin{subfigure}{0.29\textwidth}
\includegraphics[width=\linewidth]{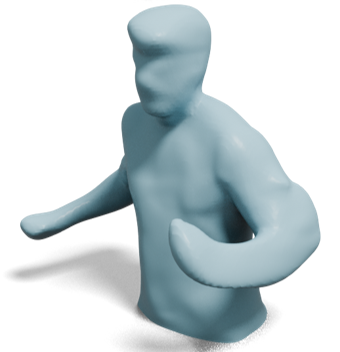}
\caption*{(c)} 
\end{subfigure}
\par\vspace{0pt} 

\end{minipage}
\caption{Qualitative comparisons of geometry field reconstruction on sequences without topological changes among (a) DynamicFusion~\cite{DBLP:conf/cvpr/NewcombeFS15}, (b) NDR~\cite{Cai2022NDR} and (c) 4DRecons.}
\label{Figure:Experiment:Compare_RGBD_w/o_topo}
\end{figure*}

\subsection{Deformation Regularization Term}
\label{Subsec:Def:Reg:Term}

The deformation regularization term $l_{\deform}(\theta) = l_{\deform}^1(\theta) +l_{\deform}^2(\theta) $ has two components. The first component $l_{\deform}^1(\theta)$ enforces that the underlying deformation between adjacent frames $t \sim \bs{f}^{\theta}(\cdot, t)$ and $t+1 \sim \bs{f}^{\theta}(\cdot,t+1)$ is as rigid as possible (ARAP)~\cite{huang2021arapreg, geolatent}. In $l_{\deform}^1(\theta)$, we want to enforce that the color field is consistent between adjacent frames. The second component $l_{\deform}^2(\theta)$ ensures that the deformations across $\bs{f}^{\theta}(\cdot, t-1)$, $\bs{f}^{\theta}(\cdot,t+1)$, and $\bs{f}^{\theta}(\cdot,t+1)$ are smooth. 

To formulate the deformation prior and enforce the consistency of the color field, we need to solve the fundamental problem of computing correspondences between the implicit fields. 4DRecons builds on the approach in GenCorres~\cite{DBLP:journals/corr/abs-2304-10523} for computing correspondences between adjacent implicit surfaces. While the approach of GenCorres focuses on geometric shapes, 4DRecons non-trivially extends it to include color information for correspondence computation. 

Specifically, we first apply the Marching Cube algorithm~\cite{Lorensen:1987:MC} to obtain a discrete mesh $\set{M}_t = (\set{V}_t,\set{E}_t)$ from the implicit surface $s^{\theta}(\bs{x}, t) = 0$ (See Figure~\ref{Fig:Implicit:Discretization} (Left)). Each vertex $v_{ti} \in \set{V}_t$ has a position $\bs{v}_{ti}$ and a color $\bs{c}_{ti}$, and they depend on the network parameters $\theta$.

For each $\bs{v}_{ti}$, our goal is to compute its correspondence $\bs{v}_{ti}^{+}$ and $\bs{v}_{ti}^{-}$ on the surface of frame $t+1$ and frame $t-1$, respectively. For simplicity, we describe the procedure for $\bs{v}_{ti}^{+}$ and that of $\bs{v}_{ti}^{-}$ to be identical. In the following, we use $\bs{v}_t(\bs{v}_t^{+})$ to stack the vertex positions $\bs{v}_{ti}(\bs{v}_{ti}^{+})$ into vectors. 

The implicit representation in frame $t+1$ provides one constraint on $\bs{v}_{ti}^{+}$:
\begin{equation}
\frac{\partial s^{\theta}(\bs{v}_{ti}, t)}{\partial t} + \frac{\partial s^{\theta}(\bs{v}_{ti}, t)}{\partial \bs{x}}^T(\bs{v}_{ti}^{+}-\bs{v}_{ti}) = 0.
\label{Eq:Implicit:Cons}    
\end{equation}
To uniquely determine $\bs{v}_{ti}^{+}$, we solve a global optimization problem to compute $\bs{d}_t := \bs{v}_t^{+}-\bs{v}_t$ (See Figure~\ref{Fig:Implicit:Discretization} (Right)). The objective function consists of two terms. The first term $e_{\arap}(\bs{d}_t)$ minimizes an as-rigid-as-possible energy between $\bs{v}_{t}^{+}$ and $\bs{v}_{t}$:
\begin{align}
\sum\limits_{v_{ti}\in \set{V}_t}\min\limits_{\bs{c}_{ti}^{+}}\sum\limits_{v_{tj}\in \set{N}_{ti}}\|(I+\bs{c}_{ti}^{+}\times)(\bs{v}_{ti}-\bs{v}_{tj}) -(\bs{v}_{ti}^{+}-\bs{v}_{tj}^{+})\|^2
\label{Eq:ARAP:Obj}
\end{align}
where $\set{N}_{ti}$ collects adjacent vertices of $v_{ti}$ on $\set{M}_t$; $I+\bs{c}_{ti}\times$ is a linear approximation of the local rotation from $\bs{v}_{ti}$ to $\bs{v}_{ti}^{+}$. Based on~\cite{huang2021arapreg}, we can express 
\begin{equation}
e_{\arap}(\bs{d}_t) = {\bs{d}_t}^T L_t^{\textup{arap}} \bs{d}_t
\label{Eq:ARAP:Obj2}
\end{equation}
where the expression of $L_t^{\textup{arap}}$ is deferred to the supp. material.

Similarly, the second term minimizes the color differences between adjacent frames $t$ and $t+1$. To this end, we use a linear approximation
$$
\bs{c}^{\theta}(\bs{v}_{ti}^{+},t+1) \approx  \bs{c}^{\theta}(\bs{v}_{ti},t+1) + \frac{\partial \bs{c}^{\theta}(\bs{v}_{ti},t+1)}{\partial \bs{x}}\bs{d}_{ti}
$$
where $\bs{d}_{ti}$ is the $i$-th element of $\bs{d}_t$. We then define this term as 
\begin{align}
e_{\textup{color}}(\bs{d}_t) & := \sum\limits_{v_{ti}\in \set{V}_{t}}\|\bs{c}^{\theta}(\bs{v}_{ti},t+1)-\bs{c}^{\theta}(\bs{v}_{ti},t) + \frac{\partial \bs{c}^{\theta}(\bs{v}_{ti},t+1)}{\partial \bs{x}}\cdot \bs{d}_{ti}\|^2 \nonumber
\\
&= \bs{d}_t^T L_t^{\textup{color}} \bs{d}_t -2 \bs{b}_t^{\textup{color}}\bs{d}_t + r_t^{\textup{color}}.
\label{Eq:COLOR:Obj}
\end{align}

We solve $\bs{d}_t$ by minimizing $e_{\arap}(\bs{d}_t)+ \lambda_{\textup{color}}e_{\textup{color}}(\bs{d}_t)$ with linear constraints (\ref{Eq:Implicit:Cons}).
Introduce $L_t = L_t^{\textup{arap}} + \mu_{\textup{color}} L_t^{\textup{color}}$, where $\mu_{\textup{color}}=0.001$. Let $C_t\bs{d}_t = -F_t$ be the matrix representation of (\ref{Eq:Implicit:Cons}). We arrive at the following quadratic program with linear constraints to solve $\bs{d}_t$:
$$
\bs{d}_t = \underset{\bs{d}}{\textup{argmin}} \quad \bs{d}^TL_t\bs{d} - 2\bs{b}_t^T\bs{d} \quad s.t.\quad C_t\bs{d} = -F_t  
$$
which leads to a closed-form expression of
\begin{equation}
\bs{d}_t = L_t^{-1}\Big(\bs{b}_t - C_t^T\big(C_tL_t^{-1}C_t^T\big)^{-1}\big(C_tL_t^{-1}\bs{b}_t+F_t\big)\Big). 
\label{Eq:V:T:Plus}
\end{equation}
Using (\ref{Eq:V:T:Plus}), we define the first component of the deformation regularization term 
\begin{equation}
l_{\deform}^1(\theta):= \sum\limits_{t=1}^{N-1}\big(e_{\arap}(\bs{d}_t) + \lambda_{\textup{color}} e_{\textup{color}}(\bs{d}_t)\big)
\label{Eq:Deform:Term}
\end{equation}
where $\lambda_{\textup{color}}=1.0$. The second component $l_{\deform}^2(\theta)$ enforces that the deformations are smooth between triplets of adjacent frames. Let $\bs{c}_t^{+}$ collect the optimal solution $\bs{c}_{ti}^{+}$ in (\ref{Eq:ARAP:Obj}). Let $\bs{c}_t^{-}$ be defined accordingly. Note that $\bs{c}_t^{+}(\bs{c}_t^{-})$ are linear in $\bs{v}_t^{+}(\bs{v}_t^{-})$ and $\bs{v}_t$. 
We define
\begin{equation}
l_{\deform}^2(\theta) =  \sum\limits_{t=2}^{N-1}\big(\mu_{r} \|\bs{v}_t^{+}+\bs{v}_t^{-}\|^2 + \mu_{p} \|\bs{v}_t^{-}+\bs{v}_t^{+}-2\bs{v}_t\|^2\big)
\end{equation}

\subsection{Topology Regularization Term}
\label{Subsec:Top:Reg:Term}

The topology regularization term employs the persistent diagram (PD)~\cite{books/daglib/0025666}, a widely used topological signature, to align implicit fields defined in different time frames $t$. In essence, a PD is constructed based on a \textit{filtration} of a topological space, specifically a cubic lattice evaluated on a 3D SDF volume in this context. Following the approach outlined in~\cite{DBLP:journals/cgf/GabrielssonGSG20} we use the super-levelsets $\{\bs{x}|s^{\theta}(\bs{x},t)\geq\alpha\}$ with varying $\alpha$ to build the \textit{filtration}. This \textit{filtration} process identifies \textbf{critical levelset values} that influence changes in the surface topology of the level sets. These \textbf{critical values} signify the formation of voids or holes and their convergence into fewer solid entities (see Figure \ref{Figure:PD_2}).
Our topology regularization is designed to directly govern these \textbf{critical levelset values} and their presence in a differentiable manner. Consequently, our regularization aligns the topology of all SDF-induced super-levelsets along with their zero-levelset surfaces.

A PD consists of a set of 2D points, each of which corresponds to the birth and death times of a topological feature of the \textit{filtration}.
In our setting, the PD $PD(s^{\theta}(\cdot,t)) =\{(b_i,d_i), i\in \set{I}_t\}$ of $s^{\theta}(\cdot,t)$ is given by pairs of \textbf{local minima/maxima}  of \(s^\theta(\cdot,t)\). Therefore, we can backpropagate gradients w.r.t. PDs further to \(s^\theta(\cdot,t)\), as is done in \cite{DBLP:journals/cgf/GabrielssonGSG20}.
In the following, we use \(PD^i_t\) to denote \(\{b_i, d_i\}\) given by \(PD_t = PD(s^\theta, t)\).

For two timestamps \(t, t'\), we minimize the \(p\)-Wasserstein distance between the corresponding PDs $PD_t$ and $PD_{t'}$:
\begin{equation}
d(PD_t, PD_{t'}) = \left(\inf_{\substack{\sigma:\mathcal{I}_t\rightarrow\mathcal{I}_{t'} \\ \sigma\in S_{|\mathcal{I}_t|}}}\sum_{i\in \mathcal{I}_t} \lvert PD_t^i - PD_{t'}^{\sigma(i)} \rvert_p^p \right)^{1/p}
\label{Eq:PD:Distance}
\end{equation}
where \(\sigma\) is a permutation of indices \(\mathcal{I}_t\). We defer the illustration of aligning PDs to the supp. material.

Having established the distance of PD, we define the PD loss as a summation of the PD distances in pairs between timestamps $[1,N]$.
Here we sum PD distances over pairs on evenly spaced cycles along timestamps.
The final PD loss is defined as
\begin{equation}
    l_{\topo}(\theta)=\sum_{k=1}^3\frac{1}{\lvert C(k)\rvert}\sum_{i,j\in C(k)}d(PD_i, PD_j) 
\end{equation}
where \(C(k)\) is the collection of edges starting from 1 with \(k-1\) timestamps skipped, e.g., \(C(2) = \{(1, 3), (3, 5), \cdots\}\). Similar to training point cloud generators~\cite{DBLP:conf/cvpr/FanSG17}, minimizing (\ref{Eq:PD:Distance}) combines alternating optimization of permutation \(\sigma\) and $\theta$.

\subsection{Optimization}
\label{Subsec:Optimization}

The total objective function in (\ref{Eq:Total:Opt}) consists of multiple objective terms with very different energy landscapes. Direct end-to-end optimization can easily fall into local minimums. As shown in Figure~\ref{Figure:Overview}, we use a four-step optimization procedure that leads to an improved solution. The first step initializes $\bs{f}^{\theta}$ using the data term. The second step refines the geometry branch while ignoring regularization terms related to the color branch. The third step refines the color branch while fixing the geometry branch. The fourth step fine-tunes the network by activating all loss terms. We again defer the details to the supp. material.

\section{Experimental Evaluation}
\label{Section:Results}

This section presents an experimental evaluation of 4DRecons. We begin with the experimental setup in Section~\ref{Subsec:Experimant:setup}. Section~\ref{Subsec:Experimant:Compare_Geometry} and Section~\ref{Subsec:Experimant:Compare_Color} compare 4DRecons with state-of-the-art approaches in the geometry reconstruction and texture reconstruction. Section \ref{Subsec:Ablation:Study} presents an ablation study. Please refer to the supp. material for more results. 

\subsection{Experimental Setup}
\label{Subsec:Experimant:setup}

\noindent\textbf{Datasets.} To evaluate 4DRecons and baselines, we use a wide range of sequences from (1) DeepDeform~\cite{bozic2020deepdeform} dataset, (2)KillingFusion~\cite{DBLP:conf/cvpr/SlavchevaBCI17} dataset and (3) a dataset captured by us using an iPhone 14 Pro. In particular, the dataset captured by us emphasizes the sequences in which topological changes occur. For every sequence, we only leverage the RGB-D sequences and the camera intrinsics.

\noindent\textbf{Baseline approaches.} We compare 4DRecons with four current top-performing baseline approaches to evaluate the geometry and texture reconstruction: (1) a non-deep learning-based approach DynamicFusion~\cite{DBLP:conf/cvpr/NewcombeFS15} using our own implementation; (2) a state-of-the-art neural-based monocular dynamic reconstruction approach NDR~\cite{Cai2022NDR}; (3) D-NeRF~\cite{DBLP:conf/cvpr/PumarolaCPM21} that represents a dynamic scene with non-rigid geometries as a neural deformable volumetric function; (4) Hexplane~\cite{Cao2022FWD} that decomposes a 4D spacetime grid into six feature planes for novel view synthesis.

\begin{wraptable}{r}{9cm}
\fontsize{6pt}{7pt}\selectfont
\vspace{-0.17in}
\caption{Quantitative comparisons on the geometry and color field reconstruction in three datasets: DeepDeform (D\_D), KillingFusion (K\_F), Our collected Data (O\_D).}
\begin{tblr}{
colspec = {Q[l] Q[c] Q[c] Q[c] Q[c] Q[c] Q[c] Q[c] Q[c]},
  cell{1}{1} = {r=2}{},
  cell{1}{2} = {c=2}{},
  cell{1}{4} = {c=3}{},
  cell{1}{7} = {c=3}{},
  vlines,
  hline{1,3,7-8} = {-}{},
  hline{2} = {2-9}{},
}
Method        & Input Modality &       &Geometry (in mm.) $\downarrow$   &               &            & Color (PSNR in dB.) $\uparrow$  &      &      \\
              & RGB            & Depth & D\_D     & K\_F & O\_D & D\_D  & K\_F & O\_D \\
D-NeRF        &     \raisebox{0.6ex}{\scalebox{0.7}{$\sqrt{}$}}       &   $\odot$    &     2.891        &      3.139         &     4.912       &        28.78    &      27.73         &      22.86     \\
Hexplane      &     \raisebox{0.6ex}{\scalebox{0.7}{$\sqrt{}$}}       &   $\odot$    &       2.319        &      2.968         &      4.628      &      \textbf{32.79}      &       \textbf{31.28}        &      27.11     \\
DynamicFusion &      \scalebox{0.85}[1]{$\times$}          &   \raisebox{0.6ex}{\scalebox{0.7}{$\sqrt{}$}}    &     5.428        &      4.129         &      14.19      &      \scalebox{0.85}[1]{$\times$}      &      \scalebox{0.85}[1]{$\times$}         &      \scalebox{0.85}[1]{$\times$}    \\
NDR           &       \raisebox{0.6ex}{\scalebox{0.7}{$\sqrt{}$}}         &   \raisebox{0.6ex}{\scalebox{0.7}{$\sqrt{}$}}    &       0.923      &        1.323       &     1.899       &    31.08        &      30.92         &      25.09     \\
Ours          &       \raisebox{0.6ex}{\scalebox{0.7}{$\sqrt{}$}}         &   \raisebox{0.6ex}{\scalebox{0.7}{$\sqrt{}$}}    &       \textbf{0.884}      &       \textbf{1.249}        &       \textbf{1.823}     &      32.04     &      30.17         &        \textbf{27.72} 
\end{tblr}
\vspace{-0.1in}
\label{TAB:Quantitative}
\end{wraptable}

In Table~\ref{TAB:Quantitative}, We summarize the input modalities supported by different approaches and whether each approach is used for geometry or color comparisons. For fair comparisons, we modify the implementation of D-NeRF~\cite{DBLP:conf/cvpr/PumarolaCPM21} and Hexplane~\cite{Cao2022FWD} to take RGB-D inputs.



\noindent\textbf{Evaluation protocol.} We conduct both qualitative evaluations, where we visually compare the reconstruction results of 4DRecons and the baseline approaches, and quantitative evaluations. Quantitative evaluations report reconstruction errors in geometry by the difference between the reconstructed mesh and depth values inside the mask, as well as color by PSNR between rendering results and masked input RGB images.



\subsection{Evaluation on Geometry Reconstruction}
\label{Subsec:Experimant:Compare_Geometry}

\begin{wrapfigure}{r}{0.46\textwidth}
\vspace{-0.215in}
\begin{tabular}{@{}c@{}c@{}c@{}}

\includegraphics[width=0.15\textwidth]{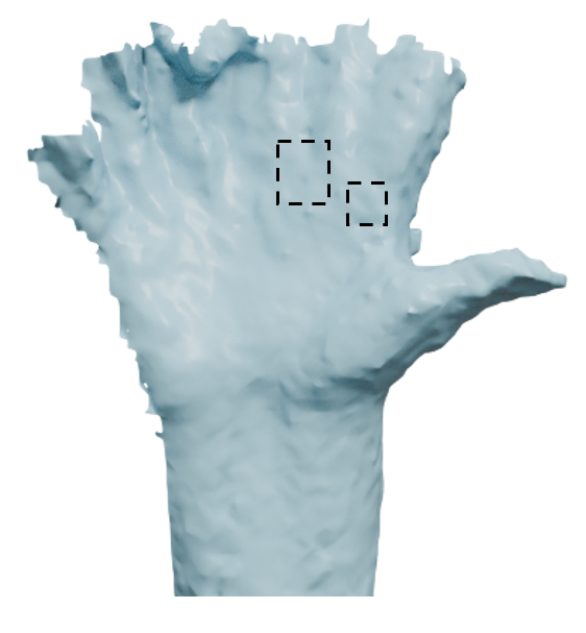}&
\includegraphics[width=0.15\textwidth]{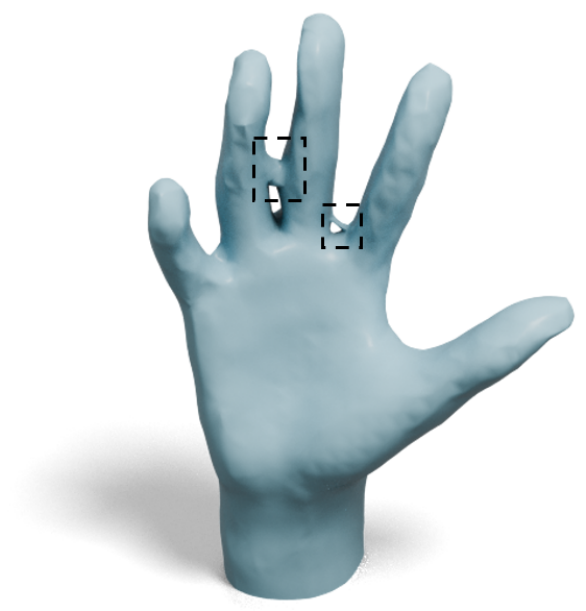}&
\includegraphics[width=0.15\textwidth]{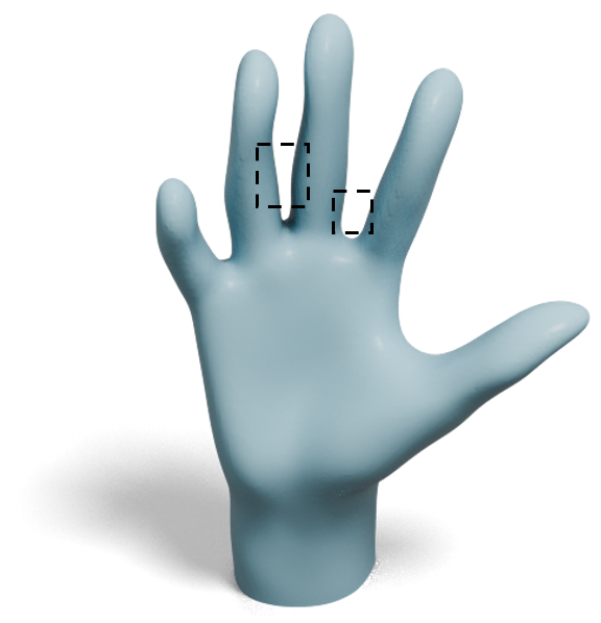}\\
\begin{minipage}{0.15\textwidth}
\centering
\footnotesize DynamicFusion
\end{minipage} &
\begin{minipage}{0.15\textwidth}
\centering
\footnotesize NDR
\end{minipage} &
\begin{minipage}{0.15\textwidth}
\centering
\footnotesize Ours
\end{minipage} \\
\end{tabular}
\vspace{-0.08in}
\caption{Qualitative evaluation on sequences with topology changes.}
\vspace{-0.15in}
\label{Figure:Experiment:Compare_RGBD_w_topo}
\end{wrapfigure}

Figure~\ref{Figure:Experiment:Compare_RGBD_w/o_topo} and Figure~\ref{Figure:Experiment:Compare_RGBD_w_topo} present qualitative results. As shown in Figure \ref{Figure:Experiment:Compare_RGBD_w/o_topo}, 4DRecons can reconstruct detailed geometries. The non-deep learning-based method DynamicFusion~\cite{DBLP:conf/cvpr/NewcombeFS15} exhibits artifacts around the surface. Moreover, the neural-based method NDR~\cite{Cai2022NDR} fails to recover the underlying approximate articulated deformation, especially in cases with large deformation and fast motion. As shown in Figure \ref{Figure:Experiment:Compare_RGBD_w_topo}, our topology regularization term \ref{Subsec:Top:Reg:Term} allows 4DRecons to enforce that the reconstruction topology remains fixed and consistent over time. In contrast, baselines produce an inconsistent and unfixed topology, as different parts of fingers randomly merge when they are close to each other.

Quantitatively, as shown in Table~\ref{TAB:Quantitative}, our approach reduces the mean reconstruction error of D-NeRF, Hexplane, DynamicFusion, and NDR by an average of 64.17\%, 60.14\%, 80.21\%, and 4.45\% on all the datasets, respectively. Quantitative improvements are consistent with qualitative results. These numbers again show the effectiveness of 4DRecons in integrating observations from different frames by modeling suitable deformation and topology regularization losses.

\subsection{Evaluation on Color Field Reconstruction}
\label{Subsec:Experimant:Compare_Color}

\begin{wrapfigure}{r}{0.61\textwidth}
\vspace{-0.18in}
\begin{tabular}{@{}c@{}c@{}c@{}}
\includegraphics[width=0.2\textwidth]{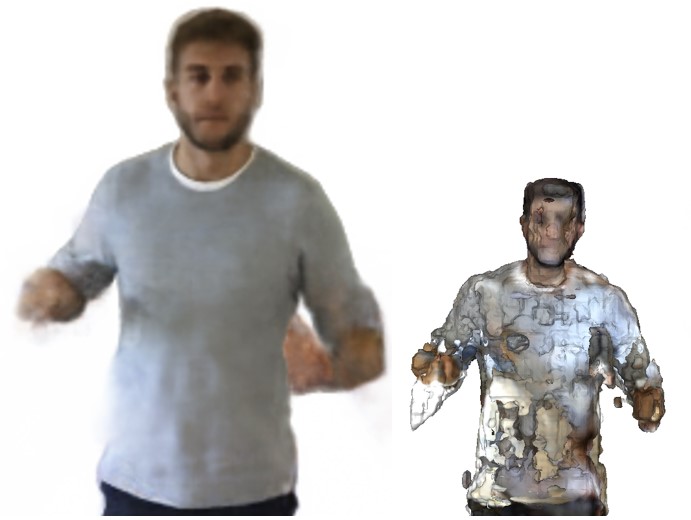}&
\includegraphics[width=0.2\textwidth]{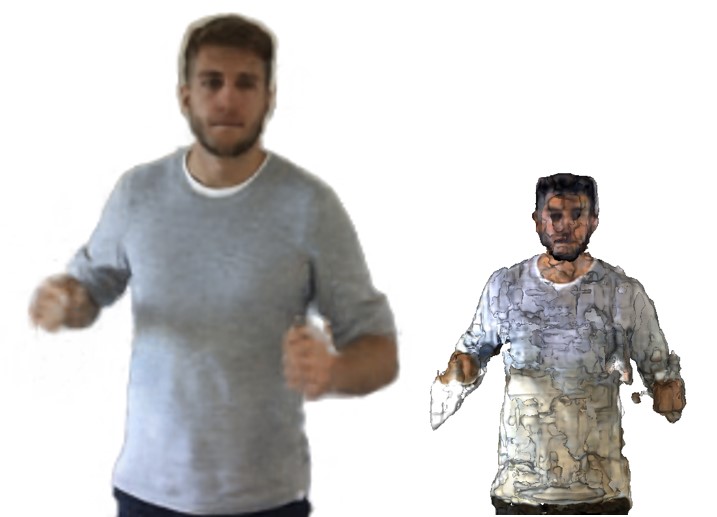}&
\includegraphics[width=0.2\textwidth]{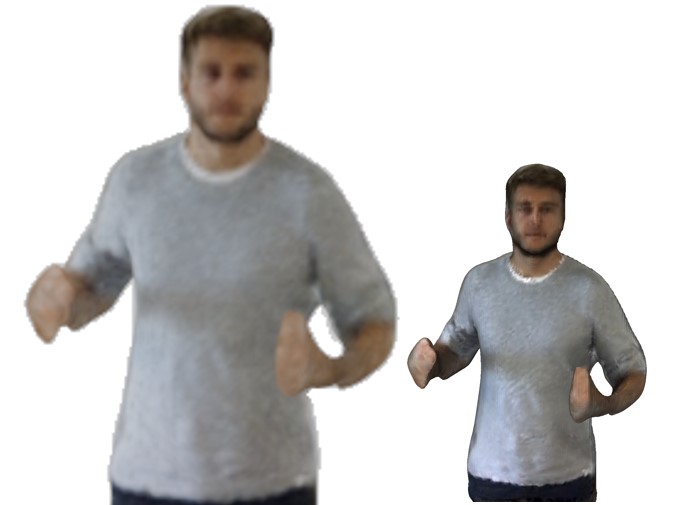}\\
\begin{minipage}{0.2\textwidth}
\centering
\footnotesize D-NeRF~\cite{DBLP:conf/cvpr/PumarolaCPM21}
\end{minipage} &
\begin{minipage}{0.2\textwidth}
\centering
\footnotesize Hexplane~\cite{Cao2022FWD}
\end{minipage} &
\begin{minipage}{0.2\textwidth}
\centering
\footnotesize Ours
\end{minipage} \\
\end{tabular}
\caption{Qualitative evaluation on the color field. The center is the rendering results, and the lower right corner is the colored mesh.}
\label{Figure:Experiment:Compare_Color}
\vspace{-0.1in}
\end{wrapfigure}

Figure~\ref{Figure:Experiment:Compare_Color} shows visual comparisons of the texture quality. 4DRecons can achieve rendering results that are on par with baselines trained via volume rendering and image-based optimization. Quantitative comparisons are consistent with qualitative results, as shown in Table~\ref{TAB:Quantitative}. However, when comparing the extracted textured mesh, 4DRecons yields significantly more detailed and sharper results than the baselines. This is particularly encouraging because NeRF-based techniques tend to overfit training data and show artifacts under novel viewpoints and poses. The benefit of reconstructing a textured mesh is that it enables fast rendering and many downstream applications.

\subsection{Ablation Study}
\label{Subsec:Ablation:Study}

\begin{wraptable}{r}{7.4cm}
\fontsize{6pt}{7pt}\selectfont
\vspace{-0.17in}
\caption{Quantitative ablation study on the geometry and color field reconstruction in three datasets: DeepDeform (D\_D), KillingFusion (K\_F), Our collected Data (O\_D).}
\vspace{-0.1in}
\begin{tblr}{
colspec = {Q[l] Q[c] Q[c] Q[c] Q[c] Q[c] Q[c]},
  cell{1}{1} = {r=2}{},
  cell{1}{2} = {c=3}{},
  cell{1}{5} = {c=3}{},
  vlines,
  hline{1,3,7-8} = {-}{},
  hline{2} = {2-7}{},
}
Model & Geometry (in mm.) $\downarrow$   &               &            & Color (PSNR in dB.) $\uparrow$  &      &      \\
      & D\_D     & K\_F & O\_D & D\_D  & K\_F & O\_D \\
w/o $l_{\deform}$     &       1.371         &      3.227         &      3.903      &     27.19       &     29.84          &     19.61     \\
w/o $l_{\topo}$    &       0.920        &       1.383        &     2.249      &      31.79      &      30.08        &      25.97       \\
w/o $e_{\textup{color}}$     &      0.947          &    1.480         &   2.421         &    30.17        &      29.22         &      22.82       \\
w/o Optim Steps     &     4.827      &     9.762          &       9.162     &      17.48      &       14.93        &       14.05     \\
Complete     &       \textbf{0.884}      &       \textbf{1.249}        &       \textbf{1.823}     &      \textbf{32.04}     &      \textbf{30.17}         &        \textbf{27.72}  
\end{tblr}
\vspace{-0.1in}
\label{TAB:Quantitative_Ablation}
\end{wraptable}

Quantitatively, we validate the effectiveness of different components of 4DRecons with an extensive ablation study in Table~\ref{TAB:Quantitative_Ablation}. Row 5 shows our complete model as a reference. In rows 1–4 we remove the four components one at a time from the complete model, observing that our optimization procedure introduced in \ref{Subsec:Optimization} (row 4) provides the largest quantitative benefit, which is essential for the model's convergence. It is followed by the deformation regularization term (row 1) defined in \ref{Subsec:Def:Reg:Term}, color consistency term (row 3) defined in \ref{Eq:COLOR:Obj}, and topology regularization term  (row 2) defined in \ref{Subsec:Top:Reg:Term}. Each of them plays its own role in enhancing the geometry and color field reconstruction.

\begin{wrapfigure}{r}{0.36\textwidth}
\vspace{-0.175in}
  \begin{center}
    \includegraphics[width=0.35\textwidth]{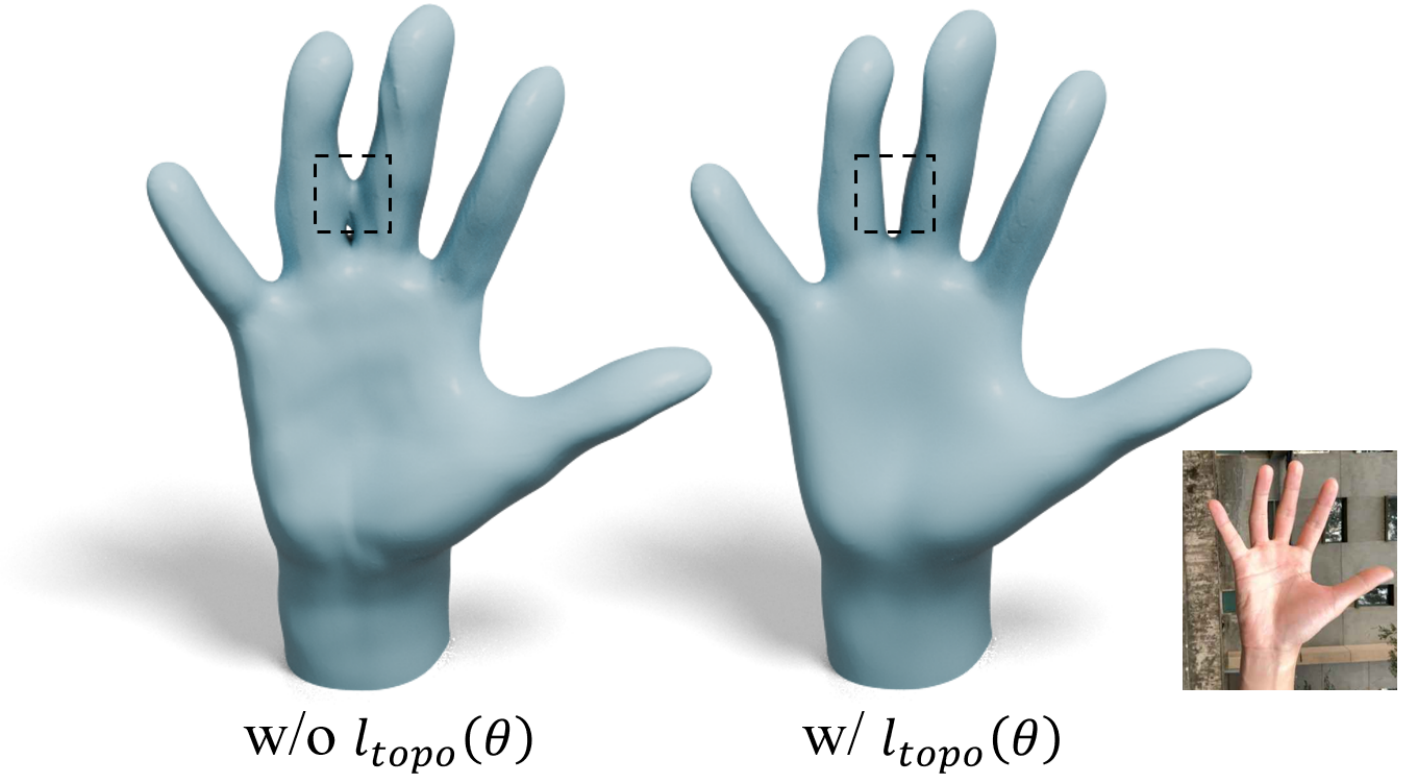}
  \end{center}
  \vspace{-0.1in}
  \caption{Comparisons of geometry field reconstruction with (right) and without (left) $l_{\topo}(\theta)$.}
  \label{Figure:Ablation:Topology}
  \vspace{-0.2in}
\end{wrapfigure}

\begin{figure}[tb]
  \centering
  \begin{subfigure}{0.4\linewidth}
    \centering
    \includegraphics[height=2.8cm]{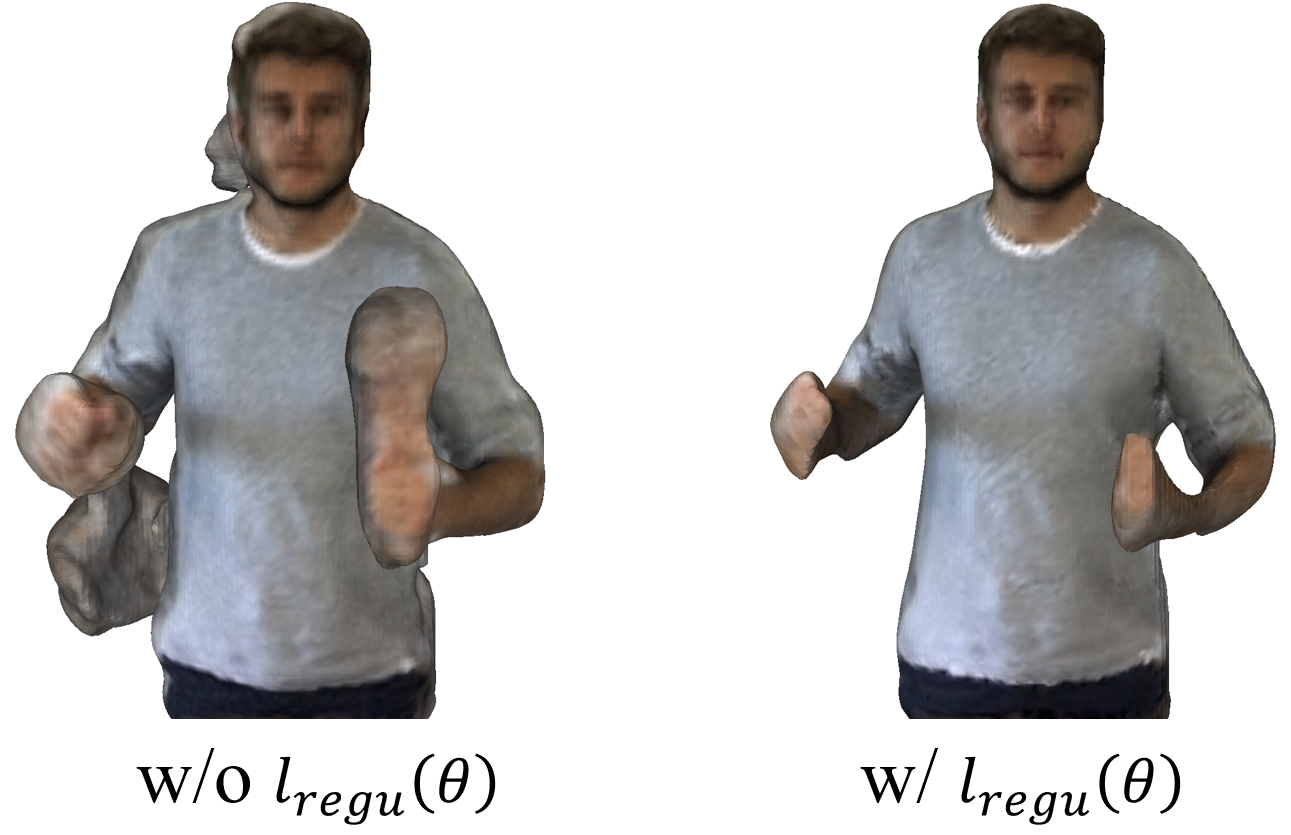}
    \caption{}
    \label{Figure:Data:Term}
  \end{subfigure}
  \hfill
  \begin{subfigure}{0.58\linewidth}
    \includegraphics[height=2.8cm]{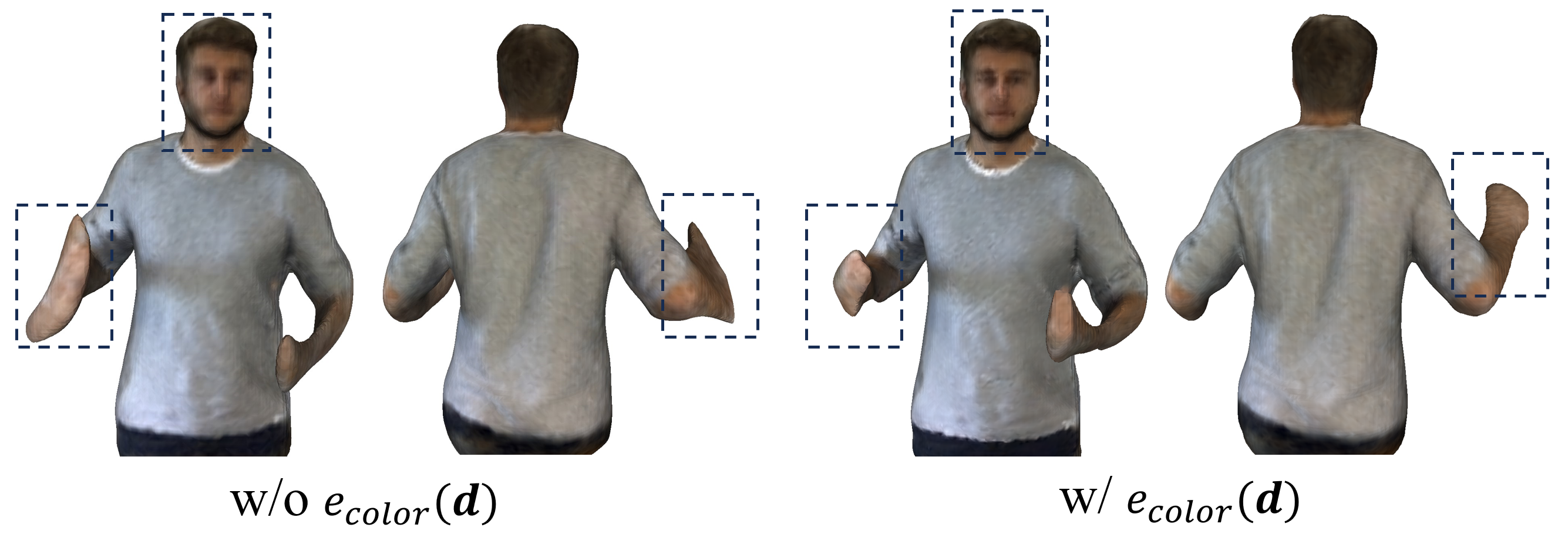}
    \caption{}
    \label{Figure:Ablation:Color:Consistency}
  \end{subfigure}
  \caption{(a) The qualitative comparisons with (right) and without (left) the regularization term $l_{\regu}(\theta)$ defined in (\ref{Eq:DataRegu:Term}). (b) The qualitative comparisons with (right) and without (left) the color consistency term $e_{\textup{color}}$ defined in (\ref{Eq:COLOR:Obj}). }
  \label{fig:short}
\end{figure}

Qualitatively, Figure~\ref{Figure:Ablation:Deformation} illustrates that the deformation regularization term can significantly improve the reconstruction results. Omitting this term leads to incomplete reconstructions with deformations that are neither smooth nor locally rigid. This is expected, as we rely on the deformation term to propagate partial observations across the entire sequence. Without this term, propagation is done by the smoothness of the network, which does not understand the underlying approximate articulated motions. Figure \ref{Figure:Ablation:Topology} indicates the topology regularization term is crucial for enforcing the topology to remain fixed by aligning the persistent diagram (PD) throughout the sequence. Dropping this term compromises the topological consistency, thus diminishing the performance of 4DRecons. Finally, as shown in Figure~\ref{Figure:Ablation:Color:Consistency}, enforcing color consistency improves the accuracy of correspondences and therefore enhances the quality of the reconstruction, which is the key difference between our approach and the formulation in GenCorres~\cite{DBLP:journals/corr/abs-2304-10523}. Without this term, both the geometry reconstruction quality and the texture reconstruction quality drop.

\section{Limitations}

One limitation is that our approach assumes that the final reconstruction is a closed deforming surface, and it requires that each point of the underlying object be observed from at least one frame. In the future, we want to address this issue by using unsigned distance fields that can model open surfaces. 
Another limitation is that the topology regularization term, which improves the topological consistency, cannot guarantee that the topology of the reconstruction remains fixed, as it is enforced in the least-square sense. In the future, we plan to address this issue by employing an explicit mesh representation and using the implicit field to ensure that the mesh reconstruction is self-collision-free.

\section{Conclusions}

In this paper, we have introduced 4DRecons, an implicit reconstruction of geometry and color from partial scans of a deforming object obtained from a single RGB-D camera. It combines a data term, a deformation regularization term, and a topology regularization term. The geometry regularization term computes inter-frame correspondences to propagate observed geometry and color signals. The topology regularization term promotes the topological consistency of reconstructions across all frames. The experimental results show that 4DRecons outperforms baseline approaches both qualitatively and quantitatively.  

{\small
\bibliographystyle{ieee_fullname}
\bibliography{main}
}
\appendix
\newpage

Supplemental materials provide more formulation details of the deformation regularization term in Section~\ref{Sec:supp_deformation_regularization}, implementation details of the network architecture in Section~\ref{Sec:supp_architecture}, and experimental details of the hyper-parameters setup in Section~\ref{Sec:supp_implement_details}. Section~\ref{Sec:supp_illustration_PDs} introduces the illustration of PD alignment. Section~\ref{Sec:supp_geometry_field} and Section~\ref{Sec:supp_color_field} show more results on the geometry field and color field reconstruction, respectively. Section~\ref{Sec:Boundary:Conf:Value} illustrates our definition of the boundary confidence value that is used in our sampling strategy. Section~\ref{Sec:supp:Optimization_Detail} describes the optimization procedure. Finally, Section~\ref{Sec:supp_discussion} discusses ample opportunities for future research.

\section{Details of Deformation Regularization Term}
\label{Sec:supp_deformation_regularization}

\subsection{Expression of $L_t^{\textup{arap}}$}
\label{Sec:supp_arap_expression}
\begin{equation}
L_t^{\textup{arap}}(\mathbf{V}(t)) = 2L \otimes I_3 - B( \mathbf{V}(t) ) D(\mathbf{V}(t)) B(\mathbf{V}(t)) ^T ,\nonumber
\label{Eq:Sup:arap}    
\end{equation}
where $L$ is the graph Laplacian of the mesh extracted from the zero-level set of the geometry field $s^{\theta}(\bs{x},t)$. $\mathbf{V}(t)\in \R^3$ are the vertices of the mesh. $B(\mathbf{V}(t)) \in \mathbb{R}^{3n \times 3n}$ is a sparse block matrix defined as 
\begin{equation}
B(\mathbf{V}(t)) = \begin{cases} 
\sum\limits_{k \in \mathcal{N}_{i}} \bs{e}_{ik}(t) \times, & \text{if } i = j \\
 \bs{e}_{ij}(t) \times, & \text{if }  j \in \mathcal{N}_{ti} \\
0, & \text{if }  \text{else} 
\end{cases} \nonumber
\label{Eq:Sup:arap_B}    
\end{equation}
where $ \bs{e}_{ij}(t) = \mathbf{V}_i(t) - \mathbf{V}_j(t)$ and $ \mathbf{V}_i(t) $ is the $i$-th vertex of $ \mathbf{V}(t) $. $D(\mathbf{V}(t)) \in \mathbb{R}^{3n \times 3n}$ is a diagonal block matrix defined as

\begin{equation}
D(\mathbf{V}(t)) = (\sum\limits_{j\in \mathcal{N}_{i}}( \| \bs{e}_{ij}(t) \|^2 I_3 - \bs{e}_{ij}(t)\bs{e}_{ij}(t)^T))^{-1} \nonumber
\label{Eq:Sup:arap_D}
\end{equation}

\subsection{Expression of $e_{\textup{color}}$}
\label{Sec:supp_color_expression}
\begin{equation}
e_{\textup{color}}(\bs{d}_t) = \bs{d}_t^T L_t^{\textup{color}} \bs{d}_t -2 \bs{b}_t^{\textup{color}}\bs{d}_t + r_t^{\textup{color}}. \nonumber
\label{Eq:Sup:color_arap}
\end{equation}
Let $H(\mathbf{V}(t)) \in \mathbb{R}^{3n \times 3n}$ is a sparse block matrix defined as 
\begin{equation}
H(\mathbf{V}(t)) = \begin{cases} 
\frac{\partial \bs{c}^{\theta}(V_i(t),t+1)}{\partial V_i(t)}, & \text{if } i = j \\
0, & \text{if }  \text{else}
\end{cases} \nonumber
\label{Eq:Sup:arap_B}    
\end{equation}
where $i,j = 0, 1, \dots, n-1$. Let $G(\mathbf{V(t)}) \in \mathbb{R}^{3n \times 1} $ is a matrix defined as
\begin{equation}
G(\mathbf{V(t)}) = [ G_0(\mathbf{V(t)}), G_1(\mathbf{V(t)}), \dots, G_{n-1}(\mathbf{V(t)}) ] ^ T. \nonumber
\end{equation}
where $ G_k(\mathbf{V(t)}) = \bs{c}^{\theta}(V_i(t),t+1) - \bs{c}^{\theta}(V_i(t),t) $, $i = 0, 1, \dots, n-1$. Then we can express $L_t^{\textup{color}} \in \mathbb{R}^{3n \times 3n}, \bs{b}_t^{\textup{color}} \in \mathbb{R}^{1 \times 3n}, r_t^{\textup{color}} \in \mathbb{R}^{1}$ in Equation \ref{Eq:Sup:color_arap} as
\begin{align}
    L_t^{\textup{color}} &= H(\mathbf{V}(t))^T H(\mathbf{V}(t)) \nonumber \\
    \bs{b}_t^{\textup{color}} &= -\frac{1}{2}G(\mathbf{V(t)})^T H(\mathbf{V}(t)) \nonumber \\
    r_t^{\textup{color}} &= G(\mathbf{V(t)})^T G(\mathbf{V(t)}) \nonumber
\end{align}

\section{Details of Network Architecture}
\label{Sec:supp_architecture}
\begin{figure}[t]
\centering
\includegraphics[width=0.5\textwidth]{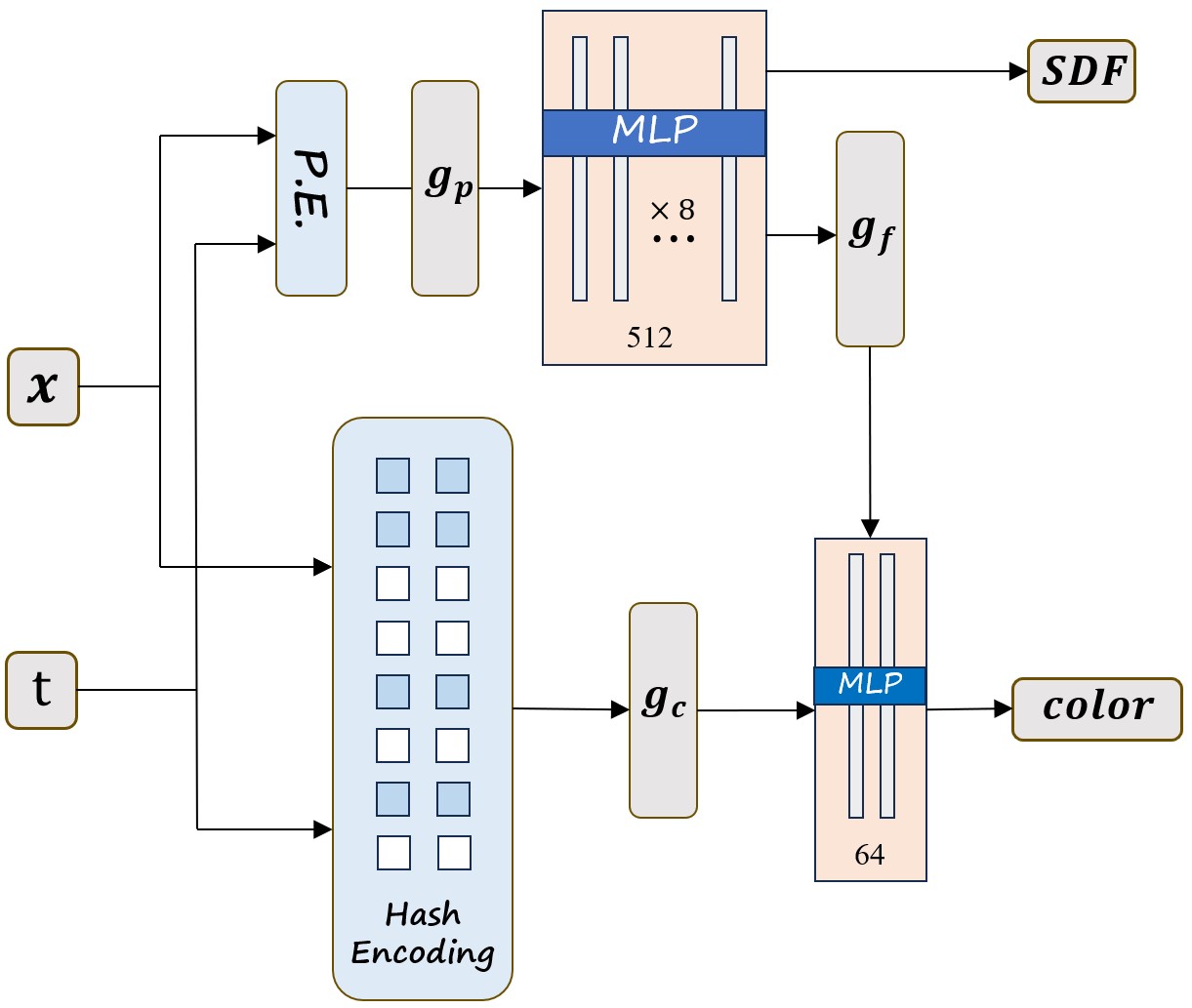}
\caption{Illustration of the network architecture.}
\label{Figure:supp_network}
\end{figure}

As shown in Figure~\ref{Figure:supp_network}, we implement our 4D implicit field $\bs{f}^{\theta}:\R^3 \times \R \rightarrow \R\times \R^3$ by combining a geometry branch $s^{\theta}:\R^3 \times \R \rightarrow \R^{1+f_g}$ and a color branch $c^{\theta}:\R^3 \times \R \times \R^{f_g} \rightarrow \R^{3}$. Both branches employ an encoder and multilayer perceptrons (MLP). Specifically, the geometry branch $s^{\theta}$ applies the positional encoding to the 3D coordinate $\bs{x}$ with 8 frequencies to derive a positional feature $\bs{g}_{p}$. We then feed the concatenation of $\bs{g}_{p}$ and the frame index $t$ into eight fully connected layers (with \textrm{softplus} activations and $512$ channels per layer) to decode the implicit value $s^{\theta}(\bs{x},t)$ and a latent feature vector $\bs{g}_{f} \in \R^{f_g}$ ($f_g = 256$). We use a skip connection ~\cite{Park_2019_CVPR} to connect the input with the output of the fourth MLP layer and weight normalization to stabilize the optimization of the parameters. The color field utilizes the multiresolution hash grid~\cite{mueller2022instant} to $\bs{x}$ and $t$ to obtain a color feature $\bs{g}_{c}$.  The number of feature dimensions per entry is $2$, the number of levels is $16$, the base resolution is $16$ and the scale factor per level is $1.3819$. We then feed $\bs{g}_{f}$ and $\bs{g}_{c}$ into three fully connected layers (using \textrm{ReLU} activations) to generate color $\bs{c}^{\theta}(\bs{x},t)$. Note that $\bs{g}_{p}$ encodes the correlation between the geometry branch and the color branch. 

\section{Implementation Details}
\label{Sec:supp_implement_details}
The deformation regularization term $l_{\deform}(\theta)$ is based on the mesh extracted from the zero-level set of the geometry field $s^{\theta}(\bs{x},t)$. We use the Marching Cube algorithm for discretization by a voxel grid of size $50 \times 50 \times 50$. The output mesh typically contains more than $5000$ vertices. Following Gencorres~\cite{DBLP:journals/corr/abs-2304-10523}, we simplify the output mesh into 2000 faces~\cite{garland1997surface} to reduce the computation complexity. The number of vertices $n$ is around $1000$.

We initialize the geometry field $s^{\theta}(\bs{x},t)$ as an approximate unit sphere~\cite{DBLP:conf/cvpr/AtzmonL20} at the beginning of training. We train our neural networks using the ADAM optimizer~\cite{journals/corr/KingmaB14}. Empirically, we set the weights in Eq. 1 as: $\lambda_{\deform}=0.001, \lambda_{\topo}=0.001$ and the weights in Eq. 4 as: $\lambda_{\sreg}=0.1, \lambda_{\creg}=0.1, \lambda_{\soff}=1.0, \lambda_{\sEDR}=1.0$. We use Fast-Robust-ICP~\cite{zhang2021fast} as a preliminary step to initialize the camera extrinsic for each frame. We use autograd in PyTorch to compute $F_t$ and $C_t$ and use fine differences to approximate other derivative computations. we use marching cubes with $128$ grid resolution to extract the zero-level set of implicit surfaces.

All the experiments are conducted on a single NVIDIA RTX A6000.

\section{Illustration of aligning PDs}
\label{Sec:supp_illustration_PDs}
\begin{figure}[h]
\centering
\includegraphics[width=0.8\textwidth]{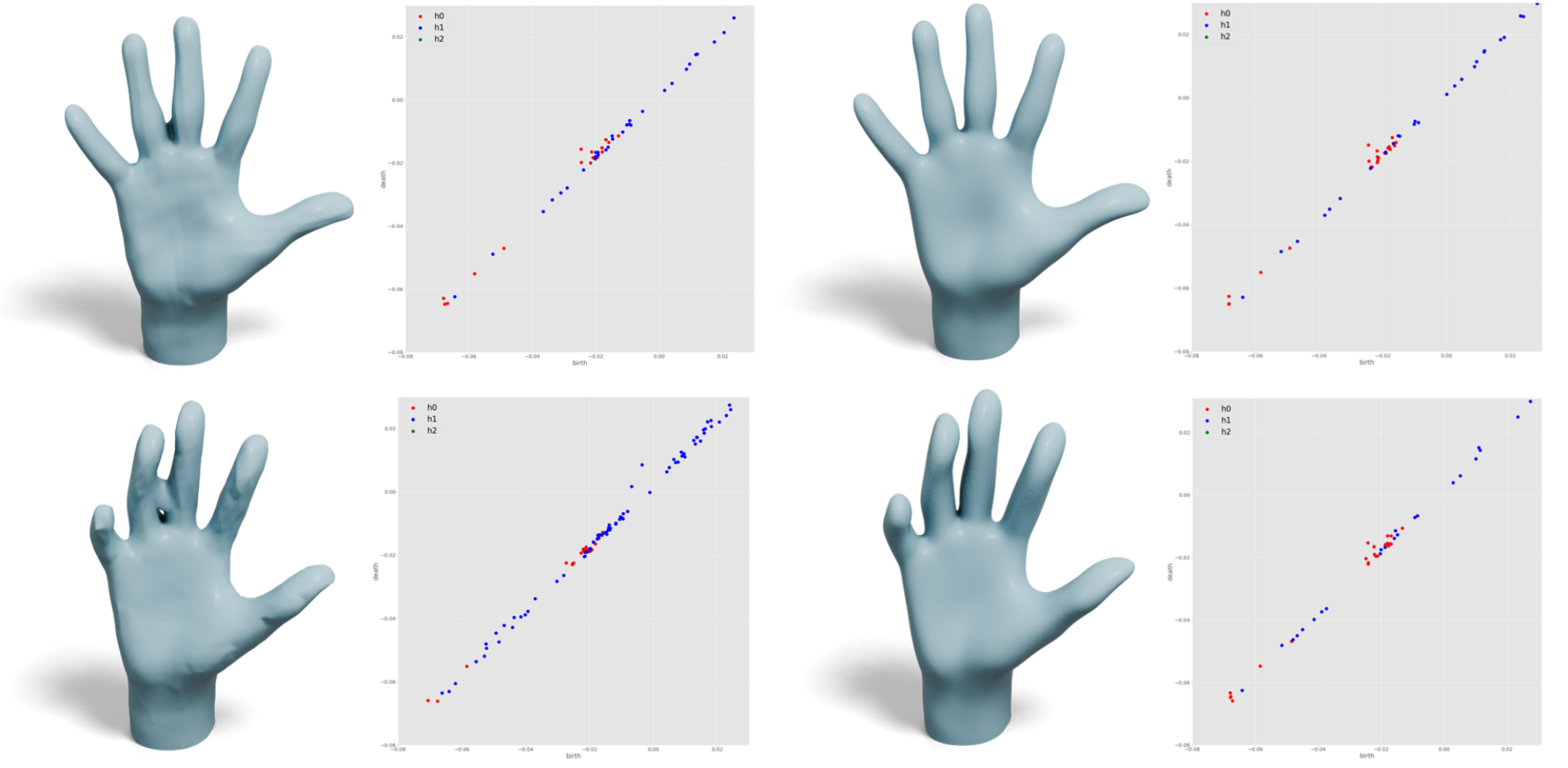}
\caption{Illustration of aligning PDs of two frames.}
\label{Figure:Supp:PD:Alignment}
\end{figure}

Our topology regularization term $l_{\topo}(\theta)$ aligns the topology of all SDF-induced super-levelsets along with their zero-levelset surfaces. As shown in Figure~\ref{Figure:Supp:PD:Alignment}, without $l_{\topo}(\theta)$, the top and bottom PDs of the left column are misaligned, indicating a lack of topological consistency. With $l_{\topo}(\theta)$, the top and bottom PDs of the right column match, showing topological consistency. 

\section{Results on Geometry Field Reconstruction}
\label{Sec:supp_geometry_field}
\begin{figure*}[t]
\centering


\begin{minipage}[b]{0.48\textwidth}
\centering
\begin{subfigure}{0.3\textwidth}
\includegraphics[width=\linewidth]{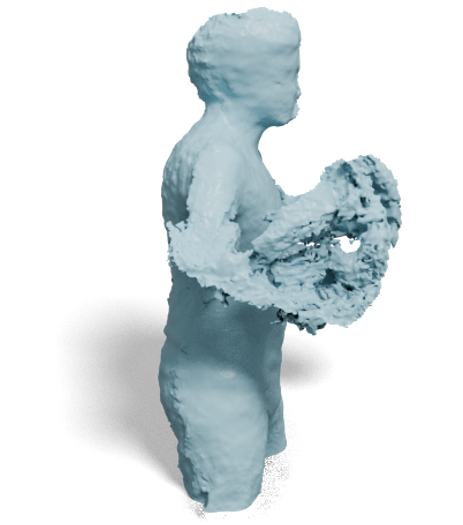}
\end{subfigure}
\hfill
\begin{subfigure}{0.31\textwidth}
\includegraphics[width=\linewidth]{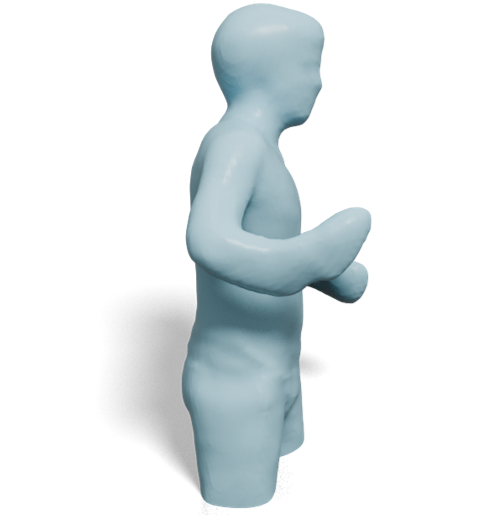}
\end{subfigure}
\hfill
\begin{subfigure}{0.3\textwidth}
\includegraphics[width=\linewidth]{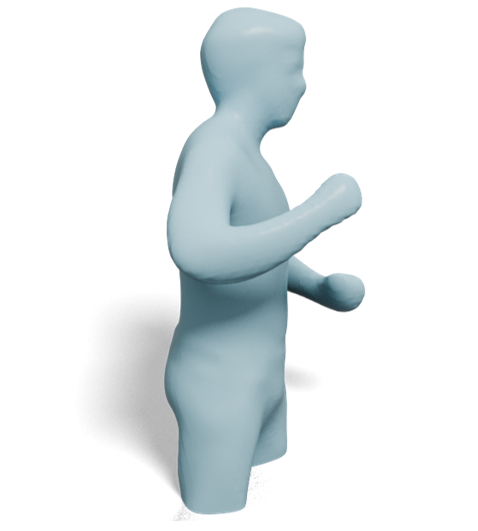}
\end{subfigure}
\par\vspace{0pt} 

\begin{subfigure}{0.3\textwidth}
\includegraphics[width=\linewidth]{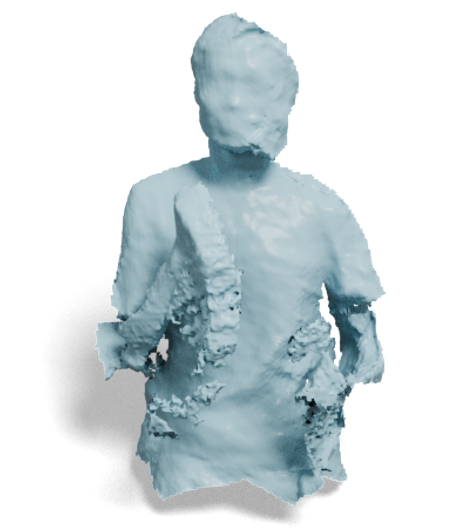}
\end{subfigure}
\hfill
\begin{subfigure}{0.3\textwidth}
\includegraphics[width=\linewidth]{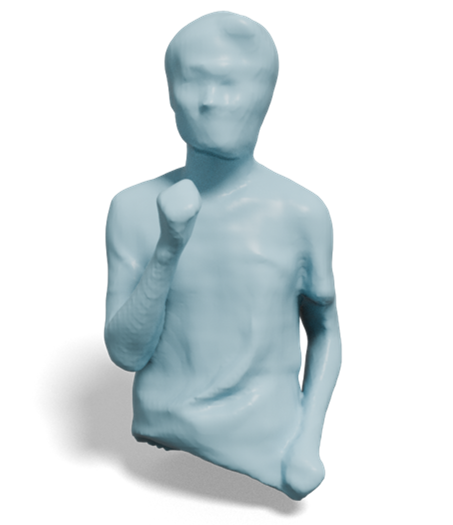}
\end{subfigure}
\hfill
\begin{subfigure}{0.3\textwidth}
\includegraphics[width=\linewidth]{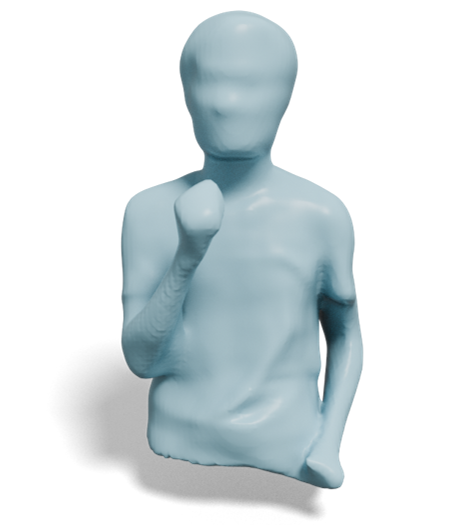}
\end{subfigure}
\par\vspace{0pt} 

\begin{subfigure}{0.3\textwidth}
\includegraphics[width=\linewidth]{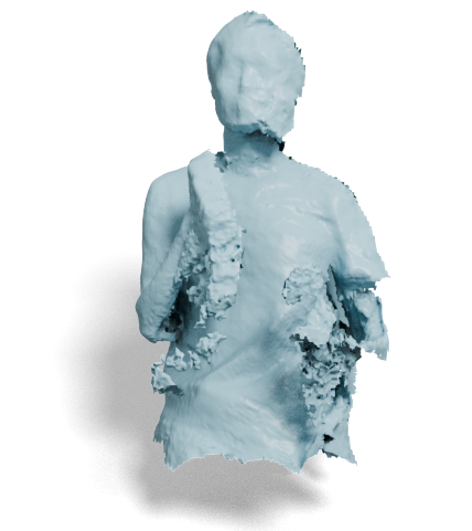}
\caption*{(a)} 
\end{subfigure}
\hfill
\begin{subfigure}{0.3\textwidth}
\includegraphics[width=\linewidth]{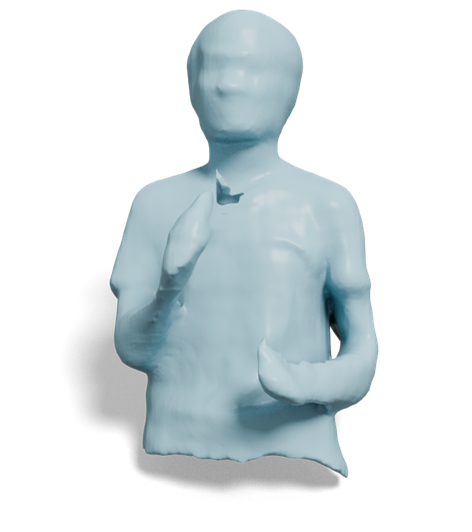}
\caption*{(b)} 
\end{subfigure}
\hfill
\begin{subfigure}{0.3\textwidth}
\includegraphics[width=\linewidth]{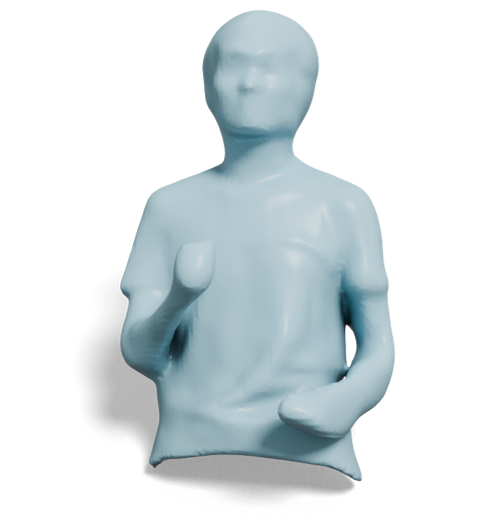}
\caption*{(c)} 
\end{subfigure}
\par\vspace{0pt} 
\end{minipage}
\hfill
\vline\hfill
\begin{minipage}[b]{0.48\textwidth}
\centering
\begin{subfigure}{0.3\textwidth}
\includegraphics[width=\linewidth]{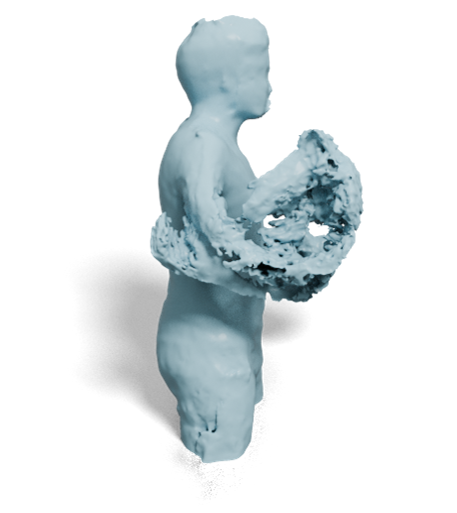}
\end{subfigure}
\hfill
\begin{subfigure}{0.3\textwidth}
\includegraphics[width=\linewidth]{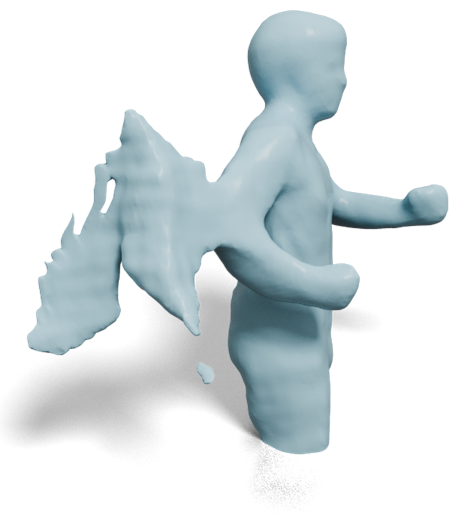}
\end{subfigure}
\hfill
\begin{subfigure}{0.3\textwidth}
\includegraphics[width=\linewidth]{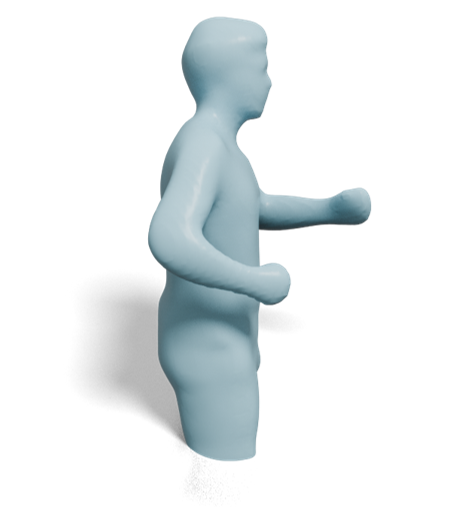}
\end{subfigure}
\par\vspace{0pt} 
\begin{subfigure}{0.29\textwidth}
\includegraphics[width=\linewidth]{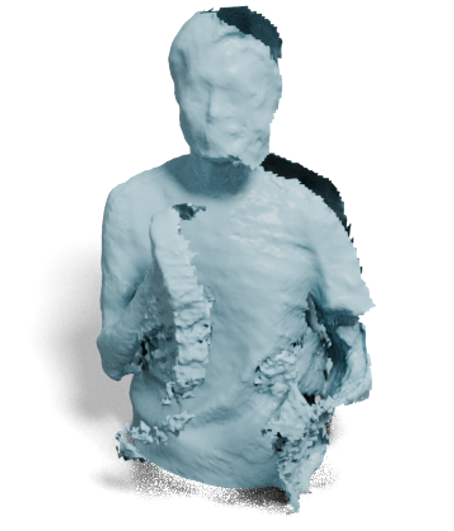}
\end{subfigure}
\hfill
\begin{subfigure}{0.3\textwidth}
\includegraphics[width=\linewidth]{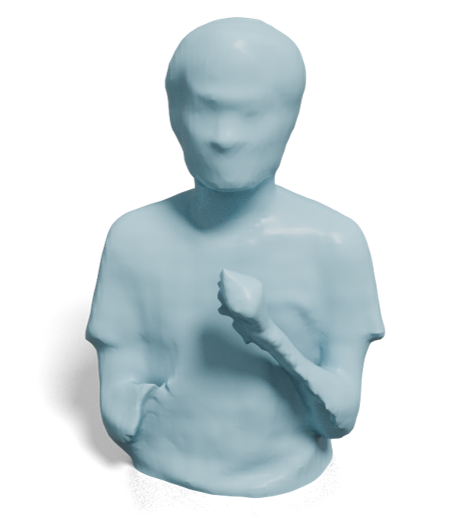}
\end{subfigure}
\hfill
\begin{subfigure}{0.3\textwidth}
\includegraphics[width=\linewidth]{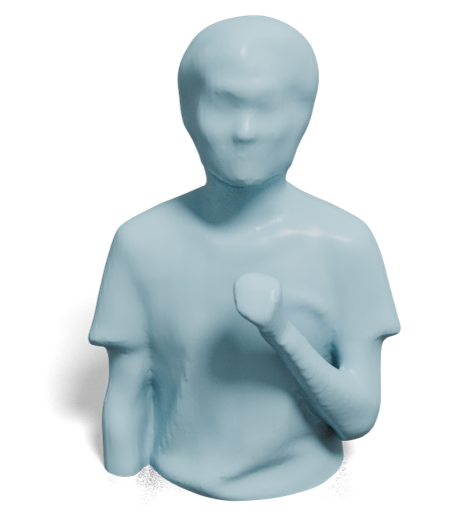}
\end{subfigure}
\par\vspace{0pt} 

\begin{subfigure}{0.3\textwidth}
\includegraphics[width=\linewidth]{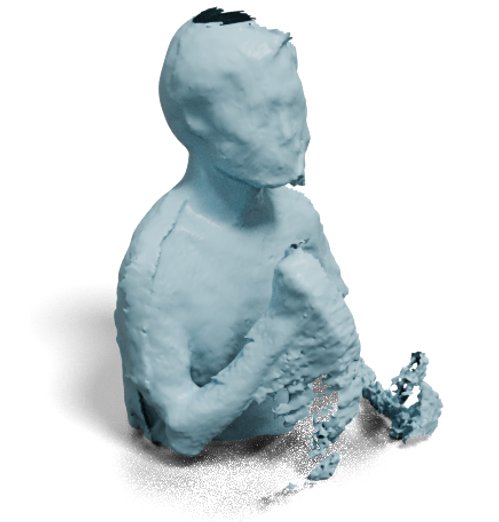}
\caption*{(a)} 
\end{subfigure}
\hfill
\begin{subfigure}{0.3\textwidth}
\includegraphics[width=\linewidth]{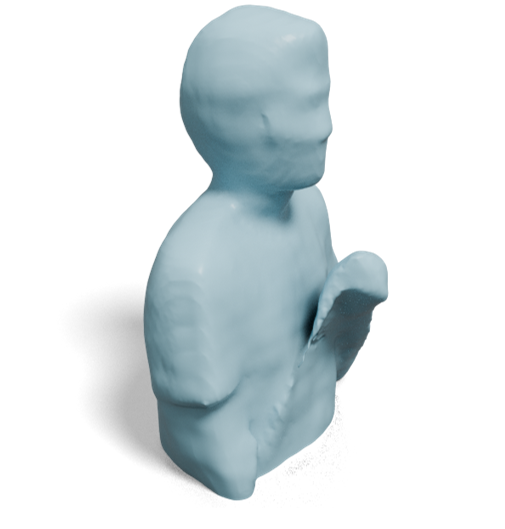}
\caption*{(b)} 
\end{subfigure}
\hfill
\begin{subfigure}{0.3\textwidth}
\includegraphics[width=\linewidth]{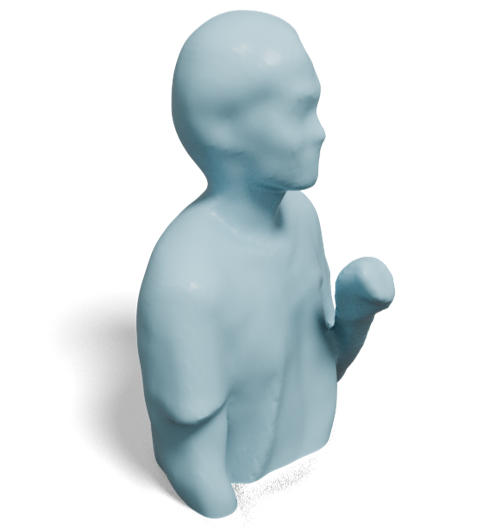}
\caption*{(c)} 
\end{subfigure}
\par\vspace{0pt} 

\end{minipage}

\caption{Qualitative comparisons of geometry field reconstruction among (a) DynamicFusion~\cite{DBLP:conf/cvpr/NewcombeFS15}, (b) NDR~\cite{Cai2022NDR} and (c) 4DRecons.}
\label{Figure:Sup_geometry}
\end{figure*}

In Figure \ref{Figure:Sup_geometry}, we show more visual comparisons of the geometry reconstruction with two important baselines DynamicFusion~\cite{DBLP:conf/cvpr/NewcombeFS15} and NDR~\cite{Cai2022NDR}. With the help of our deformation regularization term, 4DRecons can reconstruct a more complete, smooth, and detailed geometry.

\section{Results on Color Field Reconstruction}
\label{Sec:supp_color_field}

\begin{figure*}[t]
\centering


\begin{minipage}[b]{0.48\textwidth}
\centering

\begin{subfigure}{0.3\textwidth}
\includegraphics[width=\linewidth]{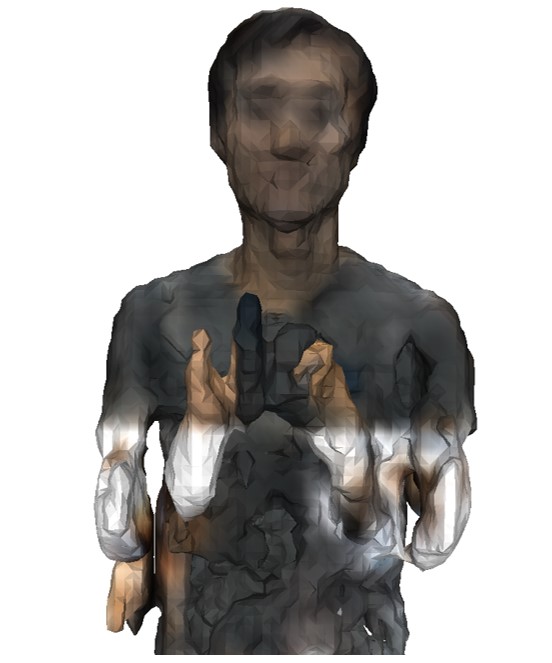}
\end{subfigure}
\hfill
\begin{subfigure}{0.3\textwidth}
\includegraphics[width=\linewidth]{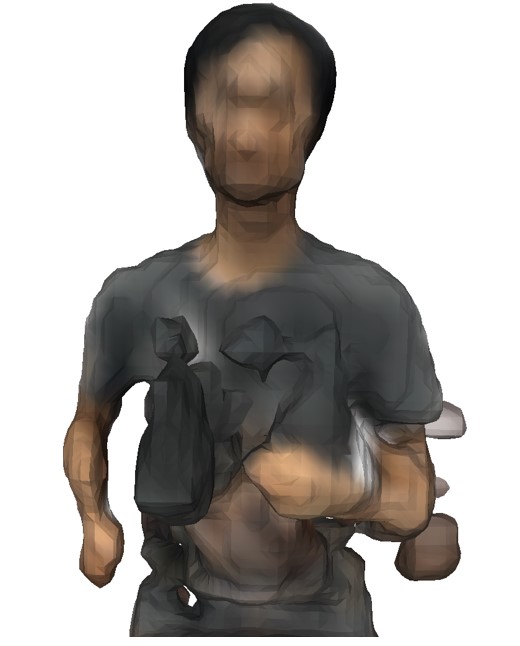}
\end{subfigure}
\hfill
\begin{subfigure}{0.3\textwidth}
\includegraphics[width=\linewidth]{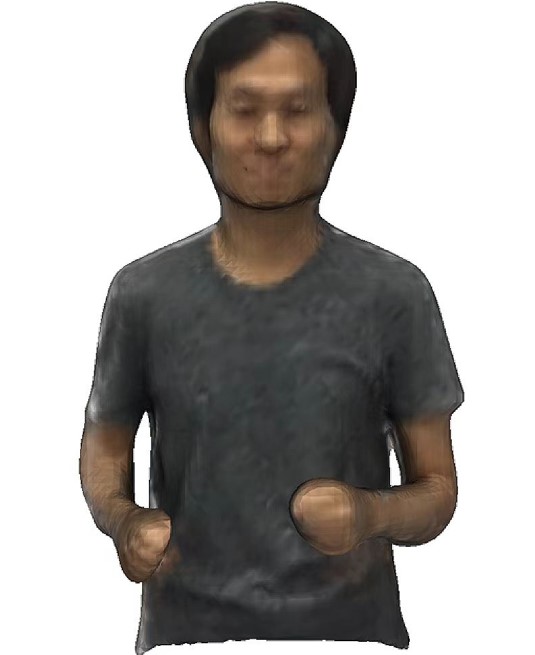}
\end{subfigure}
\par\vspace{0pt} 

\begin{subfigure}{0.3\textwidth}
\includegraphics[width=\linewidth]{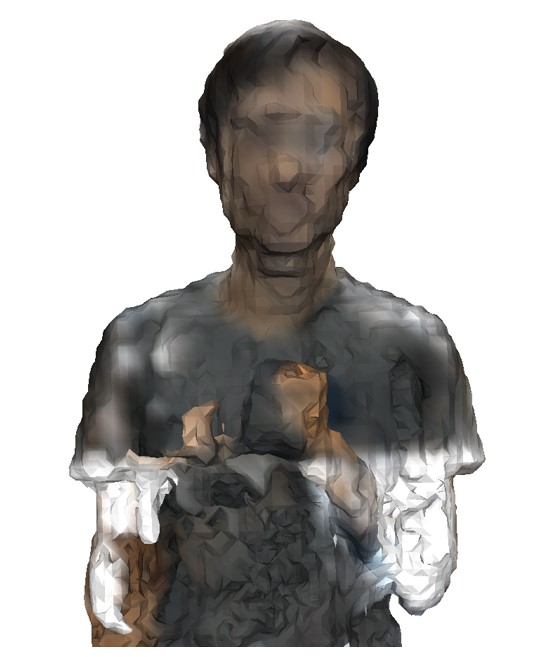}
\caption*{(a)} 
\end{subfigure}
\hfill
\begin{subfigure}{0.3\textwidth}
\includegraphics[width=\linewidth]{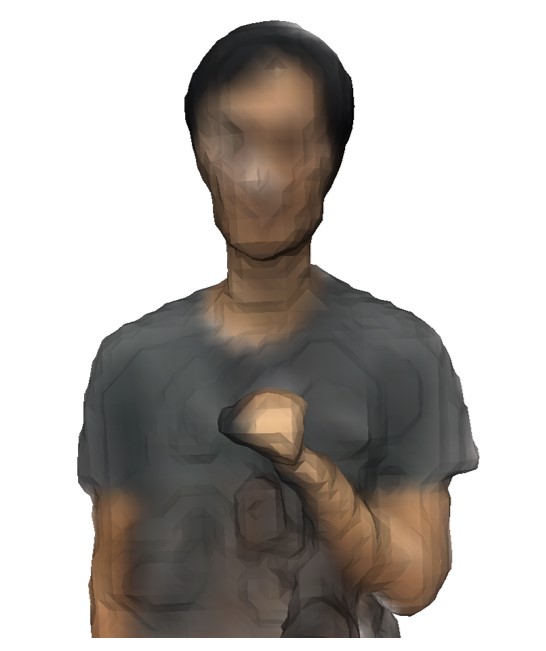}
\caption*{(b)} 
\end{subfigure}
\hfill
\begin{subfigure}{0.3\textwidth}
\includegraphics[width=\linewidth]{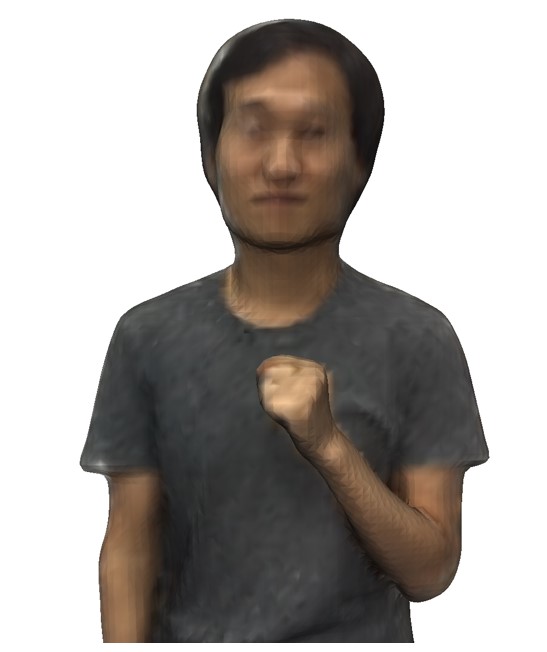}
\caption*{(c)} 
\end{subfigure}
\par\vspace{0pt} 
\end{minipage}
\hfill
\vline\hfill
\begin{minipage}[b]{0.48\textwidth}
\centering
\begin{subfigure}{0.3\textwidth}
\includegraphics[width=\linewidth]{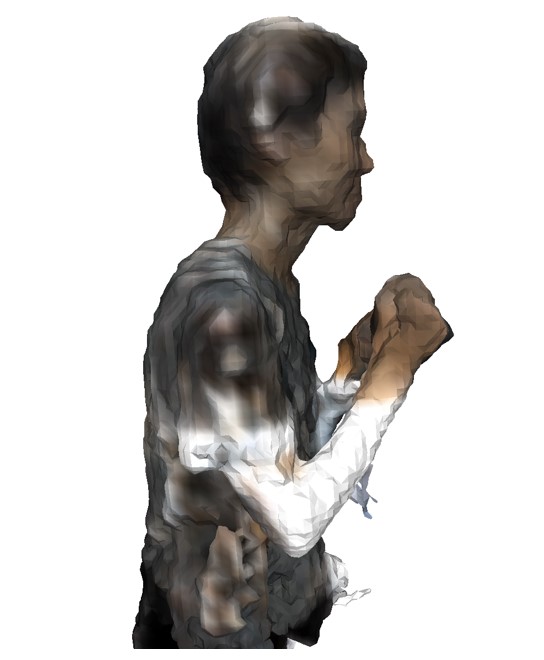}
\end{subfigure}
\hfill
\begin{subfigure}{0.3\textwidth}
\includegraphics[width=\linewidth]{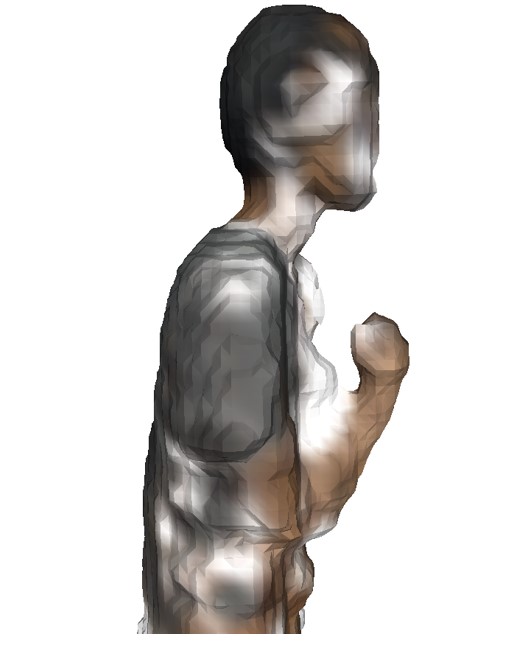}
\end{subfigure}
\hfill
\begin{subfigure}{0.3\textwidth}
\includegraphics[width=\linewidth]{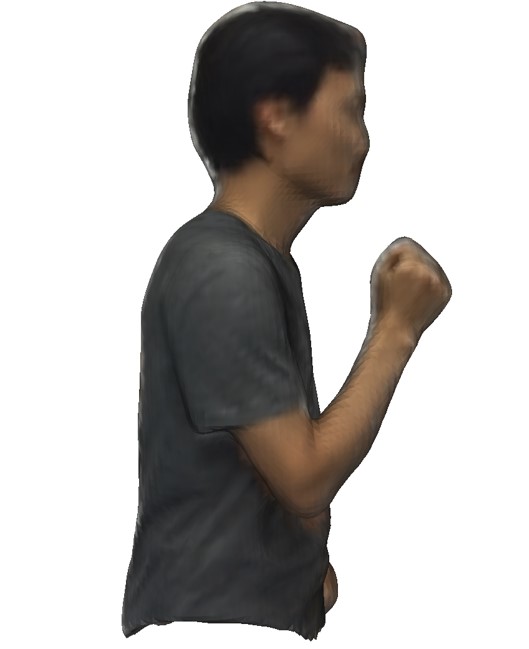}
\end{subfigure}
\par\vspace{0pt} 

\begin{subfigure}{0.3\textwidth}
\includegraphics[width=\linewidth]{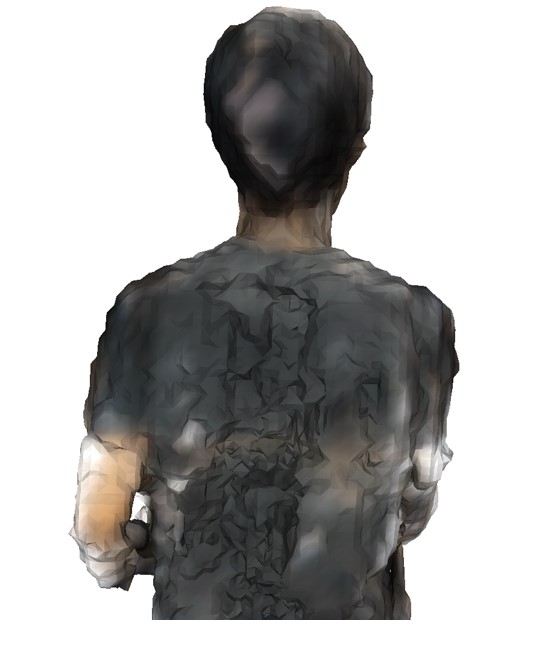}
\caption*{(a)} 
\end{subfigure}
\hfill
\begin{subfigure}{0.3\textwidth}
\includegraphics[width=\linewidth]{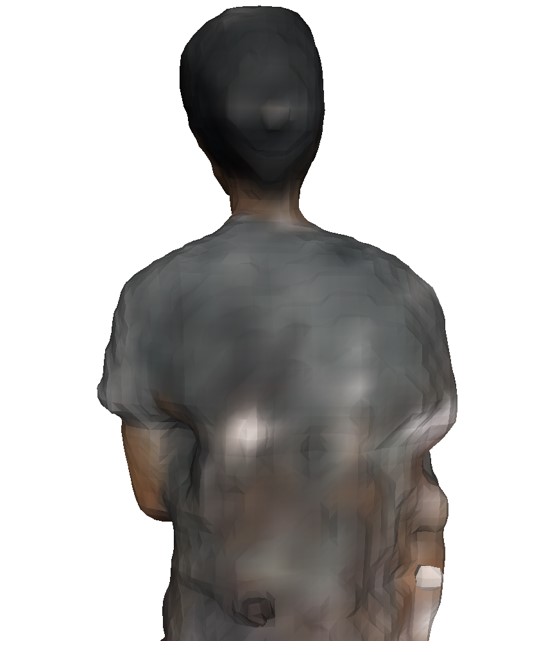}
\caption*{(b)} 
\end{subfigure}
\hfill
\begin{subfigure}{0.3\textwidth}
\includegraphics[width=\linewidth]{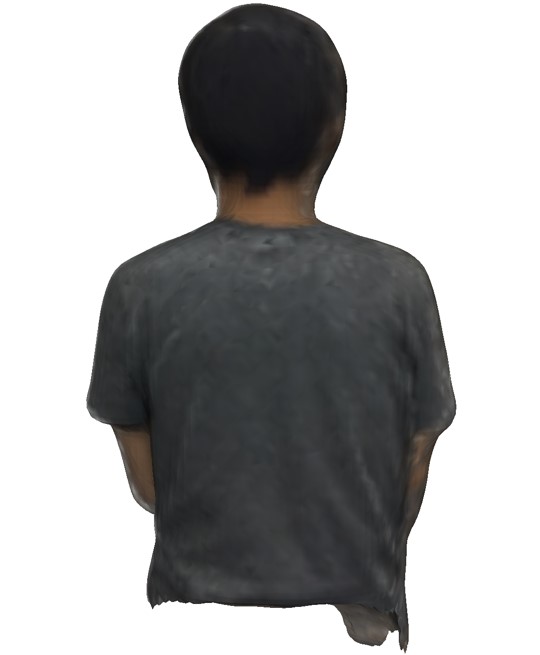}
\caption*{(c)} 
\end{subfigure}
\par\vspace{0pt} 

\end{minipage}

\caption{Qualitative comparisons of color field reconstruction among (a) D-NeRF~\cite{DBLP:conf/cvpr/PumarolaCPM21}, (b) Hexplane~\cite{Cao2022FWD} and (c) 4DRecons.}
\label{Figure:Sup_color}
\end{figure*}

In Figure \ref{Figure:Sup_color}, we present more qualitative comparisons of the color field with two important baselines, D-NeRF~\cite{DBLP:conf/cvpr/PumarolaCPM21} and Hexplane~\cite{Cao2022FWD}. 4DRecons can achieve rendering results that are on par with, but reconstruct much more detailed and sharper textured meshes than baselines trained via an imaged-based optimization procedure.


\section{Boundary Confidence Value}
\label{Sec:Boundary:Conf:Value}

We calculate the interior confidence value $b_{ti}$ by computing $\set{N}_i$, which collects the indices from $k'+1$ to $k+k'$-nearest neighbors of $\bs{p}_{ti}$ ($k = 100, k'=20$ in our experiments). For each $\bs{p}_{tj}, j\in \set{N}_i$, we project it onto the tangent plane at $\bs{p}_{ti}$ and calculate its angle $\alpha_{i,j}$ with respect to an arbitrary coordinate system in this tangent plane. Suppose that $\alpha_{i,j}$ is ordered in an increasing manner. Define
$\delta_i$ as the maximum gap between adjacent angles. Intuitively, $\delta_i$ is large if $\bs{p}_{ti}$
is close to the boundary. Note that we exclude the $k'$-nearest neighbor in the consideration to make the results robust.

\begin{figure}[h]
\centering
\includegraphics[width=0.4\textwidth]{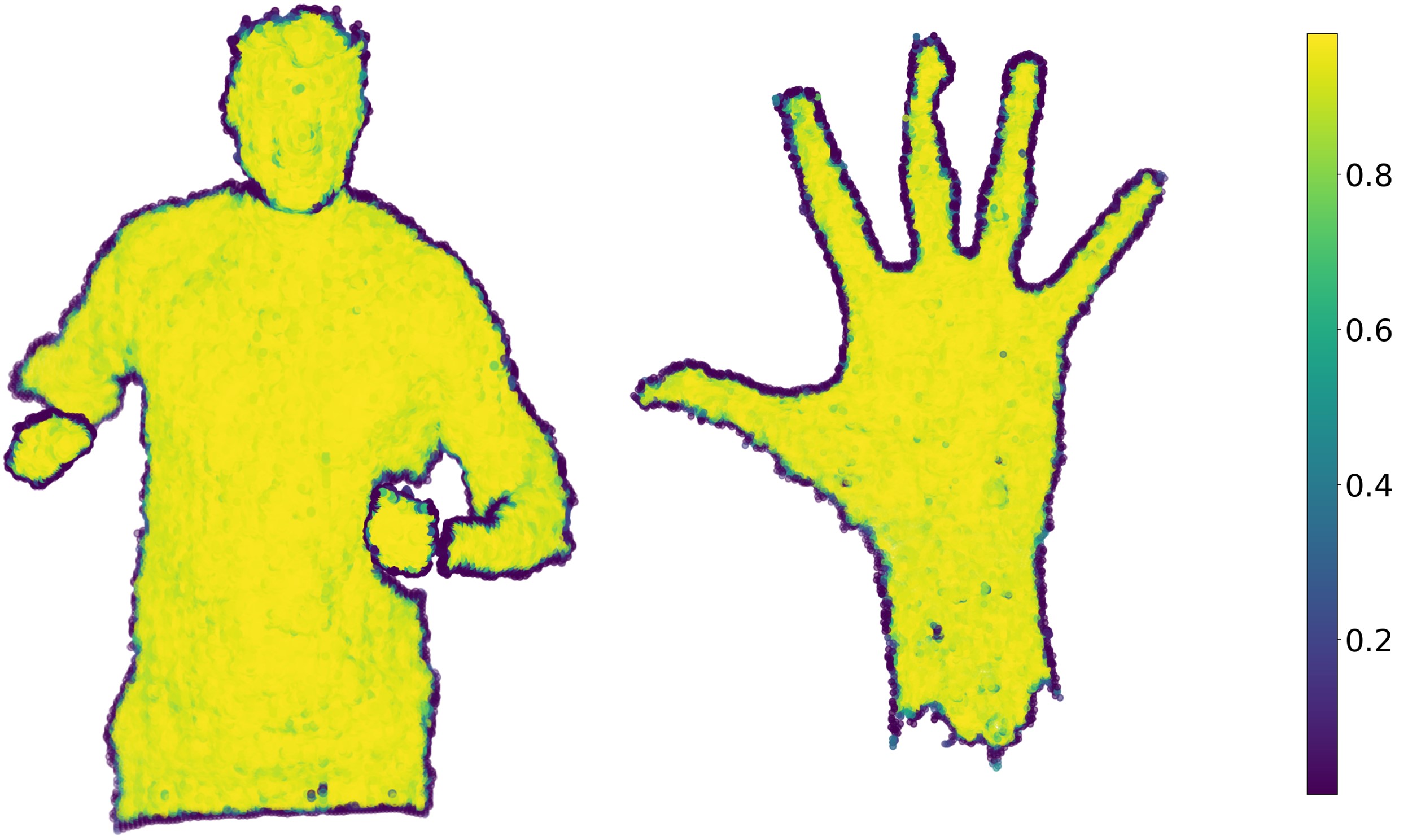}
\caption{Examples of interior confidence values on two partial scans of one dynamic sequence. In the colorbar, a lower value indicates lower interior confidence.}
\label{Figure:Boundary:Detection}
\end{figure}

We then define the interior confidence value as
$
b_{ti} = \exp\big(-\frac{\delta_i^2}{2\sigma_{a}^2}\big).
$
where we set $\sigma = \pi/4$ in this paper. Figure~\ref{Figure:Boundary:Detection} shows example results of this simple strategy. We can see that all boundary points are nicely detected. Although some interior points are also classified as boundary points, they are isolated and do not impact the overall result of regressing the implicit field.

\section{Optimization}
\label{Sec:supp:Optimization_Detail}
We employ a four-step optimization procedure to learn $\bs{f}^{\theta}$ from the input scans.
\noindent\textbf{Step I: Initialization.} The first step initializes the 4D implicit field from the input partial observations without using the regularization terms. In other words, we set $\lambda_{\deform} = 0$ and $\lambda_{\topo} = 0$ in Eq. 1. 

\noindent\textbf{Step II: Geometry field optimization.} 
The second step initializes the geometry field while ignoring the color field. In this step, we involve only the terms associated with the geometry field. Specifically, in this step we set $\lambda_{\creg} =0$ in Eq. 4 and $\mu_{\textup{color}} = 0$ in Eq. 14. 

\noindent\textbf{Step III: Color field initialization.} The third step initializes the color field while fixing the geometry field. In other words, we freeze the parameters of the geometry field by stopping the gradient backpropagation and focus on optimizing the color field with related loss terms. This step essentially uses the inter-frame correspondences derived from the geometry field and minimizes Eq. 12 to propagate color information across invisible regions at each time step.

\noindent\textbf{Step IV: Joint refinement.} In the fourth step, we jointly refine the geometry field and the color field. This allows us to use color information to obtain improved inter-frame correspondences which lead to an improved geometry field. Similarly, the improved geometry field can better propagate color information among invisible regions, resulting in more consistent and sharper texture reconstructions. 

\section{Discussion}
\label{Sec:supp_discussion}
There are ample opportunities for future research. First, 4DRecons focuses on a single sequence; it would be interesting to study how to learn a 4D representation from multiple sequences that is generalizable to new sequences. Another direction is to build a multiresolution deformation model to capture detailed deformations introduced by cloth. Finally, it is interesting to combine the strength of implicit representations and explicit representations for dynamic reconstruction. For example, one approach is to use the implicit field to guide the fusion of point clouds acquired at different frames. Potential negative societal impact: our approach requires extensive computational resources and optimization time, which may raise concerns about energy consumption. We will continue working on optimizing that in future work.

\end{document}